\documentclass[mnsc,nonblindrev]{informs3}

\OneAndAHalfSpacedXI 

\usepackage{natbib}
 \bibpunct[, ]{(}{)}{,}{a}{}{,}%
 \def\bibfont{\small}%

\usepackage[ruled] {algorithm2e}

\usepackage[center]{caption}
\usepackage{amsmath,amssymb,bm}
\usepackage{wrapfig}
\usepackage{booktabs}
\usepackage{subfigure}
\usepackage{graphicx}

\newcommand{\E}{{\mathbb{E}}}

\newcommand{\bn}{\begin{eqnarray}}
\newcommand{\en}{\end{eqnarray}}
\newcommand{\bns}{\begin{eqnarray*}}
\newcommand{\ens}{\end{eqnarray*}}
\newcommand{\textwrap}{\parbox[t]{5.5in}}
\def \Acal{{\mathcal A}}
\def \Bcal{{\mathcal B}}
\def \Dcal{{\mathcal D}}
\def \Xcal{{\mathcal X}}
\def \Lhat{\hat L}
\def \Rhat{\hat R}
\def \rhohat{\hat \rho}
\def \rhobar{\bar \rho}
\def \Vbar{\overline V}
\def \vbar{\bar v}
\def \xtilde{\tilde x}
\def \Stilde{\tilde S}
\def \Wtilde{\tilde W}
\def \omegatilde{\tilde \omega}
\def \Omegatilde{\tilde \Omega}
\def \Fcal{{\mathcal F}}
\def \Pcal{{\mathcal P}}
\def \Fcaltilde{\tilde \Fcal}
\def \Pcaltilde{\tilde \Pcal}

\TheoremsNumberedThrough     
\ECRepeatTheorems

\EquationsNumberedThrough    

\MANUSCRIPTNO{} 

\begin{document}


\RUNAUTHOR{Wang, Nascimento and Powell}

\RUNTITLE{Dynamic Bidding for Advance Commitments}

\TITLE{ Reinforcement Learning for Dynamic Bidding in Truckload Markets: an Application to Large-Scale Fleet Management  with Advance Commitments 
}

\ARTICLEAUTHORS{%
\AUTHOR{Yingfei Wang}
\AFF{Foster School of Business, University of Washington, Seattle, WA 98195, \EMAIL{yingfei@uw.edu}}
\AUTHOR{Juliana Martins Do Nascimento}
\AFF{Department of Operations Research and Financial Engineering, Princeton University, Princeton, NJ 08544, \EMAIL{jnascime@princeton.edu}}
\AUTHOR{Warren Powell}
\AFF{Department of Operations Research and Financial Engineering, Princeton University, Princeton, NJ 08544, \EMAIL{powell@princeton.edu}} 
} 
\ABSTRACT{%
Truckload brokerages, a $\$$100 billion/year industry in the U.S., plays the critical role of matching shippers with carriers, often to move loads several days into the future.  Brokerages not only have to find companies that will agree to move a load, the brokerage often has to find a price that both the shipper and carrier will agree to.  The price not only varies by shipper and carrier, but also by the traffic lanes and other variables such as commodity type.  Brokerages have to learn about shipper and carrier response functions by offering a price and observing whether each accepts the quote.  We propose a  knowledge gradient policy with bootstrap aggregation for high-dimensional contextual settings to guide price experimentation by maximizing the value of information.  The learning policy is tested using a carefully calibrated fleet simulator that includes a stochastic lookahead policy that simulates fleet movements, as well as the stochastic modeling of driver assignments and the carrier's load commitment policies with advance booking.


}%



\maketitle

%


\section{Introduction}
Trucking is the dominant mode of freight transportation. According to a recent report by the American Trucking Association, in 2016,  trucking collected \$676.2 billion in gross freight revenues, which is almost 80 percent of the nation's freight bill and trucks carried 10.42 billion tons of freight, accounting for 70.6 percent of domestic freight tonnage. However, the process of matching shippers with carriers in the large truckload segment is surprisingly manual, often involving a human on one or both ends of a transaction to confirm acceptance of a load. 

However, the relative importance of the trucking industry is not reflected in its use of data analytics to advance the efficiency and performance of its business operations. For example, a major business model of many trucking companies is the use of Call Operations Centers. The process of matching shippers with carriers in the large truckload segment is surprisingly manual, often involving a human on one or both ends of a transaction to confirm acceptance of a load. 

\subsection{Reinforcement Learning for Dynamic Bidding in Spot Markets}
Spot market contracts are short term contracts that serve unfilled or urgent shipment delivery.   While the largest shippers have worked out contracts with the larger carriers, a substantial number of truckload movements are negotiated on the spot market, where a shipper and carrier has to come to an agreement on price for an individual load.  A freight broker serves as an intermediary between a shipper who has goods to transport and a truckload motor carrier who has capacity to move that freight.  
The truckload brokerage industry processes approximately $\$$100 billion in truckload revenue each year, handling the process of matching tens of thousands of shippers with over 900,000 drivers spread among 200,000 trucking companies. Although brokers, shippers, and carriers are each incentivized to achieve contract acceptance, shippers and carriers have an incentive to hide their true willingness to accept the contracted prices \citep{tsai2011valuation}. Hence the brokerage has to suggest a price in the hopes that both will agree.  The goal of the brokerage is to maximize the total revenue on all accepted loads, and while this favors higher prices, both parties have to accept the price.

Despite that the spot market  is of great economic significance, data on
load acceptances/rejections and spot prices still remain largely private between transaction parties,  leaving the  spot market pricing problem largely unexplored.    In our situation,   shippers  broadcast a list of loads  to the broker on an ongoing basis.  For each load, the broker offers a price (which can be dynamic) to both the shipper and one chosen carrier, observing either success or failure from both parties in each matchmaking attempt.  This offers opportunities for price experimentation. However,  experimentation is surprisingly rare in practice  \citep{harrison2012bayesian}. A commonly used approach for similar pricing applications is to first use historical data to estimate unknown model parameters and then choose the ``optimal'' price given estimated parameter values \citep{phillips2005pricing,talluri2006theory}. \cite{scott2016service} uses  empirical models to estimate spot prices for truckload services  and  analyze the drivers of and deterrents to freight
rejection using a transactional data set. Yet this approach does not address the trade-off between  exploiting the current optimal solutions and exploring with uncertain outcomes to learn the model parameters.   We need to acknowledge that the ``optimal'' price given our current current belief may not be truly optimal. 



We model dynamic bidding in truckload markets as a sequential decision-making problem with a centralized control, which naturally falls into the category of reinforcement learning.  A large brokerage company might handle 500 loads per day, but these are spread among thousands of traffic lanes with very uneven density.  To account for the unique feature that the context of bidding opportunities arises over a network with strong geographic differentiation,  in this paper, we propose an optimal learning approach where we balance maximizing revenues now while also better learning the response curves for each shipper and carrier, for each traffic lane. Implemented in a learning and planning framework, the proposed algorithm is one of the first applications of reinforcement learning in large-scale truckload markets.

\subsection{Dynamic Fleet Management with Advance Booking}
In order to evaluate the benefits of price experimentation and explain key operational and economic factors, our second focus is to build an accurate simulator of a large-scale fleet  that captures real-world operations at a very high level of detail. 
Ultimately, the broker system will run several  carrier fleet management systems in parallel (one for each carrier) and each carrier will send constant  snapshots of its network to the broker system. The snapshots include updated driver/load information - such as current driver location and destination, accumulated hours of service, and new loads committed from sources other the broker. That is, all the information the broker needs to run the carrier fleet management system. Given a price recommendation from our reinforcement learning model, if the carrier system accepts the load, then the broker will actually send the load to the carrier. Then, the carrier can in real life accept/reject the load. 

Our goal is to closely calibrate real-world  truckload operations, and build a simulator with load acceptance policies and driver dispatch policies,  which can be used as a planning tool for brokers to perform analyses to answer questions ranging from price experimentation, which loads to bid on, to identifying potential carriers who may accept the load. The idea is that if the fleet management system accepted the load, it is a good load and that carrier should also accept the load.  

Truckload motor carriers face the complex problem of managing fleets of drivers, observing hours of service commitments, managing driver productivity while also getting drivers home in time, while serving a highly stochastic set of loads. While the management of fleets of vehicles (and drivers) has received considerable attention for many years, largely overlooked is the problem of making commitments to loads offered in the future. All the prior models assume that we make decisions to move trailers, containers or drivers at time $t$, and then learn information about new demands at time $t+1$.   It is not unusual to find long-haul truckload carriers where 50 percent of the loads are called in four or more days in the future, long before the carrier knows where drivers will be at that time. To this end, our research addresses the problem of designing effective load acceptance policies for making these advance commitments, before the loads can be actually dispatched in the future.

We build on the fleet management model presented in \cite{SiDaGe09}, which handles driver characteristics at a high level of detail, but where loads evolve in strict time order (that is, we know about all loads moving at time $t$ before learning about loads moving at time $t+1$). In this paper, we develop a more detailed model of the load booking process, including the full booking profile which captures how far each shipper books loads in the future. We design the driver dispatch policy using approximate dynamic programming, and the load acceptance policy, which uses a stochastic lookahead model to estimate the probability that a load will actually be accepted. Once the acceptance probability is estimated, we propose a policy function approximation (PFA) where loads are accepted if the acceptance probability is over some minimum. PFAs are analytical functions that map states to actions (see \cite{powell2018unified}).

\subsection{Our Contribution}
This paper makes the following contributions:  1) We present what appears to be the first model of an advance booking problem for truckload trucking where we have to make commitments to loads 1-14 days into the future.  The load acceptance policy is based on a stochastic lookahead model, that is simulated using a stochastic base model, which is a first for this problem class.  2)  We propose, for the first time, a formal learning model for finding the price that maximizes the broker's revenue where we maximize the online value of information from a proposed price using the concept of the knowledge gradient (KG) and bootstrap aggregation.   3) We demonstrate the first asymptotic convergence theorems for the contextual knowledge gradient policies, where we address the issue of incomplete learning.  4) Using a carefully calibrated fleet simulator, we demonstrate the effectiveness of our KG policy over other bidding policies.  We believe that our work promotes the automation of the truckload brokerage industry beyond its current status.

The rest of this paper is organized as follows. Section \ref{sec:lr} reviews related literature. Section \ref{sec:pf}  establishes the contextual bandit formulation of the sequential decision making problems for freight brokerages.  In Section \ref{sec:KG}, we propose a contextual knowledge gradient policy with bootstrap aggregation for convex loss functions  and prove its asymptotic convergence. Section \ref{sec:sm} introduces a  fleet simulator which includes a base model to describe the evolution of drivers and loads, and the policies for driver dispatch and load acceptance with advance booking. In Section \ref{sec:cs}, we investigate the performance of the knowledge gradient policy  by conducting both  simulated experiments and real world online evaluations based on the fleet management simulator. 



\section{Literature Review} \label{sec:lr}

There are different lines of investigation that have various overlap with our work.
\subsection{Revenue Management}
In the realm of revenue management, there is sporadic literature that recognizes model uncertainty and focuses on price experimentation. With the goal of maximizing expected revenue, early applications of
dynamic pricing methods have mainly focused on industries
where the short-term supply is difficult to change \citep{gallego1994optimal}. Examples  include seats on airline flights, cabins on vacation cruises,
and rooms in hotels that become worthless if not sold by a specific time.   \cite{aviv2005partially} and \cite{lobo2003pricing} were the first to consider optimal pricing or optimal bidding with uncertainty in the market response where \cite{aviv2005partially} models uncertainty in the arrival rate
 in terms of a Gamma distribution. \cite{farias2010dynamic} assumes a single unknown demand parameter and use dynamic programming to  perform price experimentation with   limited inventory and  demand uncertainty. \cite{wang2014close} uses a nonparametric solution that does not rely on the known relationship
between the price and the demand rate.  A Bernoulli demand distribution is assumed by \cite{broder2012dynamic} with many other papers \citep{carvalho2005learning,besbes2009dynamic} falling into the category of estimating  unknown parameters  by classical statistical methods.  \cite{keller1999optimal} consider optimal experimentation when the demand is changing over time.  While insights can be drawn from the economics and revenue management literature, the setting of freight brokerages combines the uncertainty of demand response to the challenges of working with sparse data over networks (which is quite different from the network problems that arise in airline revenue management).  
 
 \subsection{Bayesian Optimization and Multi-armed Bandits} 
Bayesian optimization is a powerful strategy for optimizing objective functions that are expensive to evaluate \citep{mockus1994application, jones1998efficient, jones2001taxonomy,gutmann2001radial, qu2012ranking}. A number of  two-stage and sequential procedures exist for selecting the best alternative, which in our setting would be the best price offered to both shipper and carrier.   \cite{branke2007selecting}  made a thorough comparison of several fully sequential sampling procedures.  Another single-step Bayesian look-ahead policy first introduced by \cite{gupta1996bayesian} and then further studied by \cite{frazier2008knowledge} is called the ``knowledge-gradient policy" (KG).  It chooses to measure the  alternative that maximizes the single-period expected  value of  information. 

There are other policies proposed under the umbrella of multi-armed bandits  \citep{auer2002finite,bubeck2012regret, filippi2010parametric, srinivas2009gaussian,chapelle2011empirical}.  With additional context information, different belief models have been studied, including linear models \citep{chu2011contextual} and Gaussian process regression \citep{krause2011contextual}.  The existing literature mainly focuses on real-valued
functions. \cite{wangKG2016} and \cite{tesch2013expensive} were the first to study expensive function optimization with
binary feedbacks using parametric belief models. \cite{wang2016optimal} proposes the first contextual KG policy for personalized health care. Nevertheless, none of the above is directly applicable to the problems of  freight brokerage  with strong geographic differentiation and the involvement of two business parties.

\subsection{Truckload Brokerage Services}
 With the explosive development of e-commerce applications for transportation,  there is literature on the use of web-based systems for the development of brokering services \citep{luncean2014agent}, that increases the degree of automation
of business activities, e.g.  by providing (semi-)automated
support of allocating freight requests to transport resources.  \cite{antoniou2007dr} studies a  multi-agent brokerage system   that focuses on the semantic representation of  the offerings. Domain-specific semantics combining semantic web
technologies and service-oriented computing is presented by \cite{scheuermann2012ontologies}.  Yet, the existing literature mainly focuses on facilitating the connection of the owners of goods with the freight transportation
providers, including  the semantic representation of the shipments, query management, and security.  In addition, \cite{tsai2011valuation} proposes the pricing formula for base truckload options. \cite{lindsey2013online} studies online pricing and capacity sourcing for third-party logistics providers, which uses historical data to estimate  parameters  by classical statistical methods and uses optimization to solve for the optimal price.  To the best of our knowledge, none of the existing work in transportation  explicitly addresses the automated support for online pricing and dynamic bidding between business partners.

Bidding for freight over networks has been studied under the heading of {\it combinatorial auctions} since carriers place bids for combinations of lanes. Combinatorial auction has been mainly tackled using deterministic optimization. \cite{de2003combinatorial} suggests the use of an oracle to overcome the hurdle of expressing and communicating an exponential number of bids. \cite{lee2007carrier} and \cite{wang2005combinatorial}   provide methods for identifying good bundles  and efficiently computing the bids for those bundles. \cite{chang2009decision} formulates the carrier's bid generation over the materials, location, and load type dimensions, as a synergetic minimum cost flow problem by estimating the average synergy values between loads through the proposed approximation.  \cite{chen2009solving}  optimizes a combinatorial auction using integer programming and set partitioning to enable the complete set of all possible bids to be considered implicitly, without the need to enumerate an exponential number of bids.  \cite{figliozzi2007pricing} studies carrier decisions for a type of vehicle routing problems where  contracts are put up for bid sequentially over time.  Most of the  literature on  combinatorial auctions approach the problem from the perspective of  either  the shipper or the carrier. In our problem, we study the dynamic bidding problem from the perspective of freight brokers which propose bids and then observe the response of both shippers and carriers.

\subsection{Dynamic Fleet Management}
There has been an extensive literature on dynamic fleet management in the presence of uncertain demands. The earliest work in this setting was in the context of freight car distribution \citep{jordan1983stochastic}. This work motivated a lengthy series of papers on the dynamic vehicle allocation problem \citep{cheung1996algorithm, godfrey2002adaptive, topaloglu2006dynamic} which was centered on dynamic programming formulations. This line of research focused on managing fleets of vehicles (see \cite{powell1995stochastic}  and \cite{yang2004real} for extensive reviews). \cite{powell1987operational}  and \cite{powell1988maximizing} were some of the earliest papers to deal with the complexities of managing drivers. This work finally matured in \cite{SiDaGe09} which used approximate dynamic programming to optimize drivers, taking into consideration all the complex attributes of drivers (home domicile, equipment type, hours of service, single vs. team, border crossing capabilities). None of these papers deal with the problem of optimizing the advance booking of loads. All the prior models assume that we make decisions to move trailers, containers or drivers at time $t$, and then learn information about new demands at time $t+1$. The models may not satisfy all the demands, but there is no explicit decision of whether to accept to move a load, which is difficult when the load is to be moved in the future.


\cite{rivera2017anticipatory} studies a similar anticipatory freight selection problem by considering release-days. Nevertheless, the model is specified in long-haul roundtrips with known probabilities of freight arrivals. Meanwhile, it does not model drivers' attributes and dispatch policies.

\section{Model of Dynamic Bidding}\label{sec:pf}
The business process under consideration is the pricing of a load of freight. A shipper who needs to move a load contacts a freight broker to find a carrier.  The broker gives a bid to the shipper, and in the meantime, proceeds to offer the load to a carrier with the same quote. A contract  is rejected if either the shipper or carrier decline the offer.  The goal of the broker is to maximize the expected total revenue, which is assumed to be proportional to the transaction value of the contract. The challenge lies in the sequential nature of the bidding problem where loads come one by one. This offers opportunities for price experimentation, in that each   quote is made  in sequence so that the broker can learn from each individual response and use that information to pick the next bid. However,  active experimentation with prices is rare in practice.   It is common for the broker to follow familiar patterns when pricing freight, leading to the broker's uncertainty about the average acceptance rates achievable with alternative prices.

 We use discrete choice models (such as the logit family of supply/demand functions) for both the shipper and the carrier. Brokers can  learn shipper and carrier response functions by offering a price and observing whether they each accepts the price. We could maximize profits if we knew the probability that shippers and carriers would accept a particular price in each lane, but this information is simply not available. We thus adopt a Bayesian approach and use a prior distribution over model parameters to express our uncertainty about the shipper/carrier responses.
While  a common practice is to first estimate the unknown parameters from historical data and then choose the optimal price given those parameter values, this traditional approach  fails to acknowledge the fact that with little price variation in the past, the estimated model may not be the true model so that the ``optimal'' price given the model may not be even close to optimal. In our work, we consider a sequential decision-making formulation that explicitly addresses the balance between the  broker's goal of maximizing revenue (exploitation) and the desire to learn the true acceptance model of the shipper and carrier in order to gather information to make better decisions in the future (exploration).

\subsection{The Contextual Model with Discrete Priors}
For an incoming load, the broker or the learner is presented with an attribute vector (context). To be consistent with the prior literature (in particular, \citet{SiDaGe09}), we use $b$ to describe the attributes of a load which includes  origin, destination and load type,  and other load characteristics. For convenience, we associate with each successive load a time which is indexed in the order of their ``called-in" times. At each time step $n$, the learner is presented with the load information $b^n$ and  then proposes a bid $p^n$ of a certain price  for the load. We assume that we have a discretized set $\mathcal{P}$ of possible bidding prices from a given interval $[l,u]$. For any context and bid, there is an unknown underlying binomial probability of shipper acceptance and carrier acceptance. After choosing a bid $p$, the broker will receive two binary outcomes $y^{n+1}=(y^{c,n+1},y^{s,n+1}) \in \{-1,1\}^2$, with one obtained from the  carrier $y^c$ and the other from the shipper $y^s$. The ``accept'' or ``reject'' depends stochastically on the context and the bid.  We only get to see the result of the single bid.

We adopt probabilistic modeling for the unknown probability of acceptance.  Specifically, the broker assumes separate discrete choice models for carriers and shippers which we designate $f^c(b,p;\bm{\alpha})$ and $f^s(b,p;\bm{\beta})$ as the probability of responding ``accept.''  We will omit the subscripts on $b$ and $p$ for notational simplicity when the context is clear. The load attributes $b$ are given to the broker and the broker only has the ability to choose the bid $p$.   In this work, we use two logistic models, one for carrier  $f^c( b,p;\bm{\alpha}) = \sigma(\bm{\alpha}^T\bm{x}^c(b,p))$, and one for shipper $f^s(b,p;\bm{\beta}) = \sigma(\bm{\beta}^T\bm{x}^s(b,p))$ acceptance probability, where $\sigma(h) = \frac{1}{1+\exp(-h)}$,  $\bm{x}^c(b,p)$ and $\bm{x}^s(b,p)$ are the covariates constructed from load attributes $b$ and any possible bid $p$, with the specific form to be introduced  in Section \ref{Discrete}. We let $f(b,p;\bm{\theta}):=f^c(b,p;\bm{\alpha})f^s(b,p;\bm{\beta})$, with $\bm{\theta} = (\bm{\alpha},\bm{\beta})$.

If the proposed bid is accepted by both the carrier and the shipper, the broker can get a revenue proportional to the bid $p$. Otherwise, he gets nothing. The broker's challenge is to offer bidding prices to maximize his total revenue over time $\mathcal{R}(n) = \sum_{m=0}^{n-1} p^m\cdot Y^{c,m+1}(b^m, p^m)Y^{s,m+1}(b^m, p^m)$.   It can be generally assumed that $Y^c$ and $Y^s$ are independent, meaning that whether a carrier will accept a load has nothing to do with the shipper's acceptance decision at the same time. Hence, we define the broker's single-step utility function which calculates the expected value of a contract based on its price and the likelihood it will be accepted by both the shipper and the carrier: $\mathbb{E}[p\cdot Y^cY^s] = p f^c(b,p;\bm{\alpha})\cdot f^s(b,p;\bm{\beta})$. 

Following the work by \cite{harrison2012bayesian} and \cite{chen2015optimal}, we begin with  a sampled belief model in which. Before the first load arrives, the broker adopts $K$ ambient  demand and supply models from the known logistic family $f_k(b,p):=f(b,p;\bm{\theta}_k)=f^c(b,p;\bm{\alpha}_k) f^s(b,p;\bm{\beta}_k)$, specified by the parameters $\bm{\theta}_k$.  The set of $K$ candidates are fixed over the entire time horizon. We denote by  $q^0_k$ the multinomial prior probability of the unknown parameter values $\bm{\theta}$, assigned by the broker, to each candidate $f_k(b,p)$. We use $$K^n = (q^n_1,\cdots,q^n_K)$$ to denote the ``state of knowledge'' which captures the uncertainty in our system at time $n$, with $q^n_k:= P(\bm{\theta} = \bm{\theta}_k | \mathcal{D}^{n})$ as the posterior probability of the $k$th candidate given the previously observed $n$ data points $\mathcal{D}^n = \big\{\big(b^i,p^i,(y^{c,i+1},y^{s,i+1})\big)\big\}_{i=0}^{n-1}$.  After we make the next decision $p$ under current context $b$ and obtain the next observation $y^{n+1}$, by Bayes' rule, we have
\begin{eqnarray}\nonumber
q_k^{n+1} &=& P(\bm{\theta} = \bm{\theta}_k | \mathcal{D}^{n+1} )\\ \nonumber
	 &\propto&P(y^{n+1}|\bm{\theta} = \bm{\theta}_k, \mathcal{D}^n)P(\bm{\theta} = \bm{\theta}_k|\mathcal{D}^n)	\\ \nonumber
	 &=& q_k^nP(y^{c,n+1|}|\bm{\alpha} = \bm{\alpha}_k)P(y^{s,n+1|}|\bm{\beta} = \bm{\beta}_k) \\ \label{qq}
	 &=& q_k^n\sigma(y^{c,n+1}\cdot\bm{\alpha}_k^T\bm{x}^c(b,p))\sigma(y^{s,n+1}\cdot\bm{\beta}_k^T\bm{x}^s(b,p)).
\end{eqnarray}
Because of the contextual information, we then define $S^n = (K^n, b^n)$ to be the state of the system at time $n$.  A history of this process can be presented using:
$$h^n = \bigg(K^0, b^0, S^0 = (K^0, b^0), p^0, Y^1, K^1, b^1, S^1 = (K^1, b^1), p^1, Y^2,\cdots, p^{n-1}, Y^{n},K^n\bigg).$$ 

A policy $P^\pi$ is a function mapping from any  state $s$ to a bid in $[l,u]$. The goal of the broker is to find the optimal  policy that maximizes the expected total revenue: 
$$\max_{\pi} \mathbb{E}\bigg[\sum_{n=0}^{N-1}P^\pi(S^n)\cdot Y^{c,n+1}\big(S^n, P^\pi(S^n)\big)Y^{s,n+1}\big(S^n, P^\pi(S^n)\big)|S^0\bigg].$$

We use $\mathbb{P}^{\pi}_k(A)$ to denote the probability of event $A$ under any policy $P^\pi$ with the $k$-th model as the true model that generates the shipper's and carrier's responses. Formally, given an  arbitrary sequence of contextual information $b^n$,  each $\mathbb{P}^{\pi}_k$, $k = 1,\cdots,K$ is a probability measure on the outcome space $\mathcal{Y}$ whose elements are sequences of $y = (y^c, y^s)$ as
$$\mathbb{P}^{\pi}_k(y^1,\cdots, y^N) = \prod_{n=0}^{N-1}\sigma\big(y^{c,n+1}\cdot\bm{\alpha}^T_k\bm{x}^c(b^n,p^n)\big)\sigma\big(y^{s,n+1}\cdot\bm{\beta}^T_k\bm{x}^s(b^n,p^n)\big),$$
where $p^n$ is the  bid sequence under policy $P^\pi$.

Define the filtration $(\mathcal{F}^n)_{n=0}^N$ by letting $\Fcal^n$ be the sigma-algebra generated by $b^0,p^0,y^1,\cdots,b^{n-1},p^{n-1},y^n$. We use $\Fcal^n$ and $\mathcal{D}^n$ interchangeably. 
From the definition of the posterior distribution in Eq. \eqref{qq}, it can be shown that the posterior probabilities $\{q_k^{n}\}$ form a nonnegative submartingale under $\mathbb{P}_k^{\pi}$ and form a nonnegative supermartingale under  $\mathbb{P}_j^{\pi}$, $j \neq k$. With the fact that the posterior probabilities $q_k^{n}$  are bounded in $[0,1]$,  the martingale convergence theorem can be used to establish that the  posterior belief converges almost surely.
\begin{proposition}[Convergence of Beliefs] For any policy $P^\pi$, the posterior beliefs $\{q^{n}_k\}$ converge to a limit belief $\{q^{\infty}_k\}$, almost surely, under any probability measure $\mathbb{P}^{\pi}_i$.
\end{proposition}


\section{Bidding Policy: Knowledge Gradient with Bootstrap Aggregation} \label{sec:KG}
The knowledge gradient (KG) policy was first proposed for offline context-free ranking and selection problems with the goal of finding the best alternative (where we do not pay attention to losses while we are learning). It maximizes the value of information from a decision. In ranking and section problems,  the performance of each alternative is represented by a (non-parametric) lookup table model.  KG has been extended to various belief models (e.g.  hierarchical belief model in \cite{mes2011hierarchical}, linear belief model in \cite{negoescu2011knowledge},  logistic regression in \cite{wangKG2016}).   Experimental studies have shown good performance of the KG  policy in  learning settings with expensive experiments \citep{wang2015nested, wangKG2016, russo2014learning, ryzhov2012knowledge}. 

In particular, \cite{chen2015optimal} proposed a context-free knowledge gradient policy with  sampled belief model assuming Gaussian measurement noise. This approach handles nonlinear belief models $f(x;\theta)$ by assuming that $\theta \in \{\theta_1, \ldots, \theta_K\}$, but requires that one of the elements of this set be the true value of $\theta$.  This can be a reasonable approximation in one or two dimensions, but tends to become quite poor in three or more dimensions. \cite{he2018optimal} relaxes the restriction of a fixed set of $\theta$'s by adaptively resampling a large pool and choosing new values based on mean squared error.  While this logic was shown to be useful for higher dimensional problems (e.g. it was tested for 10 and 20 dimensional settings), it could not be used in our application setting which will require models with hundreds of parameters.


As of this writing, most of the KG variants are designed for context-free cases (e.g. effectively assuming that all the loads, including traffic lanes and commodity types, are homogeneous). \cite{wang2016optimal} formally defined the contextual bandits by a two-step Bellman's recursion and derived the  first knowledge gradient type policy for Bayesian contextual bandits. The context-specific best action, which requires finding the best function, is dramatically more difficult than finding the best single action required by the context-free case.

In what follows, in Section \ref{KGDP} we propose a contextual knowledge gradient policy with discrete priors in our freight brokerage scenario with Bernoulli responses from both the shipper and the carrier, to  explicitly consider the heterogeneity in needs and responses of different loads over the entire network. In Section \ref{sec:ACT}, we provide the first asymptotic convergence theorem for the contextual knowledge gradient policy. In Section \ref{resampling}, we  design a resampling procedure specifically for convex loss functions using bootstrap aggregation to find the most promising parameter values, without the need to maintain or enumerate a large pool as used in  \cite{he2018optimal}.
 

\subsection{Knowledge Gradient with Discrete Priors for Contextual Bandits}\label{KGDP}
We have a fixed set of $K$ shipper/carrier acceptance models $f_k(b,p ) := f(b,p;\bm{\theta}_k) = \sigma\big(\bm{\alpha_k}^T\bm{x}^c(b,p)\big)\sigma\big(\bm{\beta_k}^T\bm{x}^s(b,p)\big)$ with  fixed parameter values $\bm{\theta}_k$. We assume  in this section that one of the $K$ models is the underlying true acceptance model, but  is not known to the learner.   For each load with attributes $b$, after the broker places a bid $p \in \{p_1,\cdots,p_M\}$, he receives two binary outcomes $y=(y^c,y^s)$, with one obtained from the carrier's true Bernoulli distribution with the mean $\sigma\big((\bm{\alpha}^*)^T\bm{x}^c(b,p)\big)$, and the other one from the shipper with the mean $\sigma\big((\bm{\beta}^*)^T\bm{x}^s(b,p)\big)$.   

The knowledge gradient with contextual information is the expected improvement resulting from evaluating alternative $p$ given a context $b$ \citep{frazier2008knowledge,wang2016optimal}:
 \begin{definition}
 The knowledge gradient of placing a bid $p$ while in state $s$ with context $b$ is:
 $$\nu_{p|b}^{\text{KG}}(s) := \mathbb{E}_{\bm{\theta}}\mathbb{E}_{Y|\bm{\theta}}\big[V\big((T(s, p, Y),b)\big) - V(s)|s\big].$$
 \end{definition}

 In the case of discrete priors, the states $s$ can be represented as the pair of belief state and the context $b$, $s =\big((q_1,\cdots, q_K),b\big)$. The $T(s, p, Y)$ is the one-step transition function defined by Eq. \eqref{qq}, such that $K^{n+1} = T(S^{n},p^n, Y^{n+1})$. In our dynamic bidding problem, the value at any state $s$ is the largest expected revenue under current state:
 \small{
 \begin{eqnarray*}
 V(s) = \max_{p} \mathbb{E}[p\cdot Y^cY^s|s,p]  &=&  \max_{p} \sum_{k=1}^K\mathbb{E}[p\cdot Y^cY^s|s,p, \bm{\theta}^* = \bm{\theta}_k]q_k  \\
 &=& \max_{p} \sum_{k=1}^Kp\cdot\mathbb{E}[Y^c|s,p,\bm{\alpha}^* = \bm{\alpha}_k]\mathbb{E}[Y^s|s,p,\bm{\beta}^* = \bm{\beta}_k]q_k  \\
 &=&\max_{p} \sum_{k=1}^Kp\cdot f^{c}(b,p;\bm{\alpha_k})\cdot f^{s}(b,p;\bm{\beta}_k)q_k = \max_{p} p\sum_{k=1}^K f_k\big(b,p\big)q_k.
 \end{eqnarray*}}
 
 We use $\tilde{\sigma}_k(y^c, y^s, b, p)$ to denote the probability of generating binary responses $y^c$ and $y^s$ from the $k$th candidate model under context $b$ and price $p$. Namely, $\tilde{\sigma}_k(y^c, y^s, b, p):=\sigma(y^c\cdot\bm{\alpha}_k^T\bm{x}^c(b,p))\sigma(y^s\cdot\bm{\beta}_k^T\bm{x}^s(b,p))$. Hence at the $n$th time step, based on the transition function \eqref{qq}, we have

     {\small
   \begin{eqnarray*}
  &&\mathbb{E}_{\theta}\mathbb{E}_{Y|\theta}\big[V\big((T(S^n, p, Y),b)\big) |S^n\big] \\
  &=& \sum_{l=1}^{K}\Big\{\sum_{y^c}\sum_{y^s}\big[\max_{p'}p'\sum_{k=1}^K f_k(b,p')q_k^{n+1}\big] \tilde{\sigma}_l(y^c, y^s,b,p) \Big\}q_l^n\\
  &=& \sum_{l=1}^{K}\Big\{\sum_{y^c}\sum_{y^s}\big[\max_{p'}p'\sum_{k=1}^K f_k(b,p')\frac{ q_k^n\tilde{\sigma}_k(y^c, y^s,b,p)}{\sum_{j=1}^Kq_j^n\tilde{\sigma}_j(y^c, y^s,b,p)}\big] \tilde{\sigma}_l(y^c, y^s,b,p) \Big\}q_l^n\\
  &=& \sum_{y^c}\sum_{y^s} \sum_{l=1}^{K}\big[\max_{p'}p'\sum_{k=1}^K f_k(b,p')\frac{ q_k^n\tilde{\sigma}_k(y^c, y^s,b,p)}{\sum_{j=1}^Kq_j^n\tilde{\sigma}_j(y^c, y^s,b,p)}\big]\tilde{\sigma}_l(y^c, y^s,b,p)q_l^n\\
  &=& \sum_{y^c}\sum_{y^s} \sum_{l=1}^{K}\frac{q_l^n\tilde{\sigma}_l(y^c, y^s,b,p) }{\sum_{j=1}^Kq_j^n\tilde{\sigma}_j(y^c, y^s,b,p)}\big[\max_{p'}p'\sum_{k=1}^K f_k(b,p')q_k^n\sigma(y^c\cdot\bm{\alpha}_k^T\bm{x}^c(b,p))\sigma(y^s\cdot\bm{\beta}_k^T\bm{x}^s(b,p))\big] \\
 & = &\sum_{y^c}\sum_{y^s} \max_{p'}p'\sum_{k=1}^Kq_k^n f_k(b,p') \cdot \sigma(y^c\cdot\bm{\alpha}_k^T\bm{x}^c(b,p))\sigma(y^s\cdot\bm{\beta}_k^T\bm{x}^s(b,p)).
 \end{eqnarray*}
 }
Thus, the knowledge gradient value can be expressed explicitly as:
{\small  \begin{equation}\nu_{p|b}^{KG}(S^n) =\sum_{y^c}\sum_{y^s}\max_{p'}p'\sum_{k=1}^K q_k^n f(b,p';\bm{\theta}_k) \cdot \sigma\big(y^c\cdot\bm{\alpha}_k^T\bm{x}^c(b,p)\big)\sigma\big(y^s\cdot\bm{\beta}_k^T\bm{x}^s(b,p)\big) - \max_{p}p \sum_{k=1}^K f(b,p;\bm{\theta}_k)q_k^n. \label{KGO}
\end{equation}} The knowledge gradient policy for cumulative reward  is a policy that balances the one that appears to be the current best and  that learns the most by bidding $p$ given context $b$ at time $n$ as: 
\begin{eqnarray}\nonumber
\mathbb{P}^{KG,n}_b(S^n) &=& \arg\max_{p}p\cdot P(y^s=1, y^c=1|S^n, b, p)+\tau\nu_{p|b}^{KG}(S^n)\\  \label{PKG}
&=& \arg\max_{p}p\sum_{k=1}^Kf(b,p;\bm{\theta}_k)q_k^n +\tau\nu_{p|b}^{KG}(S^n),
\end{eqnarray}
where $\tau$ reflects a planning horizon that captures the value of the information we have gained on future loads. If the planning horizon $N$ is known beforehand, $\tau$ can be chosen as $N - n$ \citep{ryzhov2012knowledge}. Otherwise, $\tau$ can also be treated as a non-negative tunable parameter. 
\subsection{Theoretical Properties of the Contextual Knowledge Gradient Policy}\label{sec:ACT}
To study the convergence of our policy, we need to consider the possibility of {\it incomplete learning}, in which case we can never identify the true candidate model. For the context-free case, we  first identify the beliefs (that are neither zero or one) under which an  {\it uninstructive bid} (providing no value of information) will be selected due to its high function value under that belief. This will result in an unchanged posterior belief, leading to the case of {\it incomplete learning}. We finally prove that the KG policy for contextual Bernoulli bandits is asymptotically consistent based on an argument of the non-existence of such confounding beliefs for contextual cases.

  We begin by providing theoretical insights for a simpler {\it context-free} case with $K= 2$  candidate models, $f^c_k(p) = \sigma(\alpha_{k,0} + \alpha_{k,1}p)$ and $f^s_k(p) = \sigma(\beta_{k,0} + \beta_{k,1}p)$, with $\alpha_{k,1}>0$, $ \beta_{k,1}<0$ for $k = 1,2$.  We first define the {\it uninstructive bid} $\hat{p}$ such that $f^c_1(\hat{p})  = f^c_2(\hat{p})$ and $f^s_1(\hat{p})  = f^s_2(\hat{p})$. The price $\hat{p}$ is called uninstructive since an observation $y^c, y^s$ of this price, either from the carrier or the shipper,  provides no information on distinguishing the two candidate models. This causes a failure in learning of the posterior distribution as given by Eq. \eqref{qq}, resulting in $q_k^{n+1} = q_k^n$ for any $k$. It is obvious to show that there exists at most one uninstructive price $\hat{p}$ for any choice of the candidate models. 

We next examine the cases when the KG policy \eqref{PKG} charges the {\it uninstructive bid} $\hat{p}$, leading to a so-called incomplete learning where  the limiting belief $q_k^{\infty} = \hat{q}_k$ that is neither zero or one, almost surely, meaning that it can never learn which candidate is the true model. We assume throughout that $\hat{p}$ falls in the feasible price range $[l,u]$ since if the two candidate models do not intersect within the feasible region,  the arguments to follow can be used to show that the incomplete learning is not an issue and the KG policy  is asymptotically optimal. Following the terminology by \cite{harrison2012bayesian}, such $\hat{s} = (\hat{q}_1,\hat{q}_2)$ is called a {\it confounding belief}  since if the KG policy recommends $\hat{p}$ at time $n$ under $q^n_k = \hat{q}_k$ and because $\hat{p}$ is uninstructive, we have $q_k^{n+1} = q_k^n =\hat{q}_k$ and the same issue repeats in all subsequent steps. The next lemma shows that the knowledge gradient value $\nu_{\hat{p}}^{\text{KG}}(s)$  of the {\it uninstructive bid} $\hat{p}$  with any belief $s = (q_1, q_2)$ is zero. For clarity, the lemma is proved for the context-free case, but it can be easily generalized to the contextual case.
\begin{lemma}
For the context-free case, under any belief state  $s = (q_1, q_2)$, there is no value of information from evaluating the uninstructive price $\hat{p}$, i.e.  $\nu_{\hat{p}}^{\text{KG}}(s) = 0$.  
\end{lemma}
\proof{Proof.}
Recall from Definition 1 that the knowledge gradient value is defined by the expected improvement in value by evaluating   the uninstructive price  $\hat{p}$.  No matter what the outcome $y^c$ or $y^s$ is, it is  indifferent between the two candidate models, resulting in no change to the posterior belief. Thus the value in the next state remains unchanged.   \Halmos
\endproof
In the meantime, the knowledge gradient value of all other prices are always non-negative. 
\begin{proposition}[Benefits of Evaluation]\label{BoE}
For the context-free case, the knowledge gradient value of evaluating any  price $p$ is nonnegative, $\nu_{p}^{KG}(s) \ge 0$, for any state $s = (q_1, \cdots, q_K)$.
\end{proposition}
It can be proved by using Jensen's inequality with  details  provided in  \ref{apx:BoE}. This proposition also holds for the contextual case.

Recall from Eq. \eqref{PKG} that the KG policy charges the price $\mathbb{P}^{KG}(s) $ that maximizes $$\mathbb{P}^{KG}(s) = \arg\max_p p\cdot P(y^s=1, y^c=1|s,  p)+\tau\nu_{p}^{KG}(s).$$ Since we learn nothing by charging the uninformative price $\hat{p}$ while the value of information of other prices are non-negative, a necessary condition for the KG policy  to charge $\hat{p}$ in the {\it context-free} case is that under the confounding belief $\hat{s}$, $\hat{p}$ is the price that achieves the highest revenue,
$$\hat{p} = \arg\max_p p \sum_{k=1}^K f_k(p)\hat{q}_k.$$ For continuous price choices from $[l,u]$, the next proposition states that under certain choices of the candidate models, such confounding belief exists, rendering in a failure in identifying which model is true.  We begin by defining  the notion of an incomplete learning set as follows.
\begin{definition}[Incomplete learning set] The incomplete learning set  $\mathcal{L}^{inc}$ is the set of tuples $(\alpha_{k,0}, \alpha_{k,1},\beta_{k,0},\beta_{k,1})$, $k = 1,2$, that satisfies $M_2 > 0, M_1 <0$ and $M_1+M_2 \ge -1/\hat{p}$,
 where  $M_k := \alpha_{k,1}(1-\sigma_k^c(\hat{p}))+\beta_{k,1}(1-\sigma_k^s(\hat{p}))$ and $\hat{p}$ is the uninstructive bid, i.e., 
\begin{eqnarray*}
\sigma(\alpha_{1,0} + \alpha_{1,1} \hat{p}) = \sigma(\alpha_{2,0} + \alpha_{2,1} \hat{p}) \text{ and } \sigma(\beta_{1,0} + \beta_{1,1} \hat{p}) &=& \sigma(\beta_{2,0} + \beta_{2,1} \hat{p}). 
\end{eqnarray*}
\end{definition}
\begin{proposition}[Existence of Confounding Belief]\label{CB}
If the chosen candidate models satisfy the  incomplete learning conditions, i.e., $(\alpha_{1,0}, \alpha_{1,1},\beta_{1,0},\beta_{1,1},\alpha_{2,0}, \alpha_{2,1},\beta_{2,0},\beta_{2,1})\in \mathcal{L}^{inc} $, there exists a confounding belief $\hat{s} = (\hat{q}_1,\hat{q}_2)$ such that the KG policy will choose the uninformative price $\hat{p}$.  
\end{proposition} The detailed proof of Proposition \ref{CB} is provided in \ref{sec:prop3}.

Now return  to the contextual case. We show next that no matter what the prior distribution is, KG will eventually find out which model is the actual model that generates the outcomes in Theorem \ref{main}.  The theorem is proved by contradiction.
We calculate the posterior probability displacement after a single step Bayesian update. As long as the limiting probability is bounded away from 0 and 1, we can show that the displacement is larger than a positive constant, which is contradictory to the convergence assumption. This argument  is based on the proof of the non-existence of the confounding beliefs for the contextual cases, because otherwise the posterior probability for the true model could become stuck at a value within the interval $(0,1)$ due to the indistinguishability between the true model and other candidates. 

We now have to consider placing a bid in the context of an attribute vector $b$, which might include origin, destination, and commodity type, which creates a finite but large set of attributes.  Now, if an uninstructive bid $\hat{p}$ is the intersection point of $f_j(b,\hat{p})$ and $f_k(b,\hat{p})$, for a different context $b'$, it is highly likely that the two functions no longer intersect  at this price $\hat{p}$. In other words, for any fixed set of alternatives, for any $p$, it is expected that under some context, the candidate functions are distinguishable and well separated. To this end,  we assume that the context is randomly generated. According to Proposition \ref{CB}, the confounding scenario only lies in a zero measure set, hence almost surely the distinguishability will push the system out of the ``unstable attractor'' resulting from local intersections between different models, and to the ``stable fixed point'': the single point distribution over the true model. The formal proof is provided in \ref{MM}.

\begin{theorem}[Asymptotic Convergence]\label{main}
Under finitely many randomly generated contextual information, the KG policy is asymptotically consistent in that the posterior distribution under the KG policy converges to a limit belief with  $q_i^{\infty} = 1$ and $q_j^{\infty} = 0$ for $\forall j \neq i$ under $\mathbb{P}^{\pi}_i$.
\end{theorem}

\subsection{Knowledge Gradient with Bootstrap Aggregation }\label{resampling}
Many real world applications require a high dimensional logistic regression model. In our dynamic bidding problem, as explained in Section \ref{Discrete},  the discrete choice model has around 400 dimensions. Other examples arise in many online recommendation systems where a logistic regression model is a natural choice for modeling the click-through rate. 
 In these cases,   it seems too naive to assume that one of the candidate values of $\bm{\theta}$ in our pre-fixed set is the truth, which may be a reasonable approximation if $\bm{\theta}$ is a one or two dimensional vector. On the other hand, after several experiments, it is likely that the probability mass is concentrated on a very small subset of the original candidates,  indicating that others are less likely to be the true model.  To this end, we propose a bootstrap aggregation method to constantly resample the candidate set so that it can adaptively find more promising parameter values. 

Before we proceed, we would like to restate the dual objectives  we encounter in this problem. First, we want to maximize the expected total revenue $\sum_{n}p^nf(b^n, p^n;\bm{\theta}^*)$ given any context sequence $\{b^n\}_{n=0}^{N-1}$. Yet we do not know the true parameter value $\bm{\theta}$ so that we need to gather information to estimate it. That is, based on observed data,   we need to find a prediction model that has low generalization error in that with high probability, it will correctly predict the value of an unseen data point. 

In what follows,  the knowledge gradient policy is used to strike a balance between recommending the best bid and the need to gather more information in order to find a better model. In the meantime, the bootstrap aggregation is used to periodically  refresh the candidate set  based on the data produced by the knowledge gradient policy as time goes by.  Bootstrapping is a statistical  practice of measuring properties of an estimator when sampling from an approximating distribution.  One standard choice for an approximating distribution is the empirical distribution function of the observed  dataset. Bagging ({\textbf B}ootstrap {\textbf {agg}}regat{\textbf {ing}}) \citep{breiman1996bagging} is a machine learning ensemble meta-algorithm designed to improve the stability and accuracy of machine learning algorithms by combining regression models or classifications learnt from  randomly generated training sets.   It  reduces variance and helps to avoid overfitting.  

If resampling is triggered at time  $n$, meaning that we re-generate the candidate set of $K$ parameter values aiming to include more promising values,   we draw on the bagging  technique \citep{breiman1996bagging} which works as follows. At time $n$, we already have a set of $n$ observations $\mathcal{D}^n$. We then generate $K$ new  datasets $\mathcal{D}^n_k$, $k = 1,\cdots, K$, each of size $n$, by sampling from $\mathcal{D}^n$ uniformly at random with replacement. Then $K$ models are fitted by using logistic regression and finding the maximum likelihood estimator $\bm{\theta}_k^n$, based on the above $K$ bootstrapped samples.   These $K$ models are fixed as the ambient candidates until the next resampling step.  In traditional bagging,  the $K$ models are  combined by averaging the output. Instead,  we treat the equal probability as the initial prior distribution of the $K$ bootstrapped models. We then make use of previously observed data $\mathcal{D}^n$  and obtain the posterior distribution by calculating the likelihood of each model, \begin{eqnarray*}\mathcal{L}(\bm{\theta}_k^n|\mathcal{D}^n) =  \prod_{i=0}^n\sigma\big(y^{c,i+1}\cdot(\bm{\alpha}^{n}_k)^T\bm{x}^{c}(b^i, p^i)\big)\sigma\big(y^{s,i+1}\cdot(\bm{\beta}_k^{n})^T\bm{x}^{s}(b^i, p^i)\big).
 \end{eqnarray*}
With resampling,  we modify our belief state to include both the candidate parameter values and the posterior probabilities, $\big((\bm{\theta}_k^n)_{k=1}^K,q_1^{n},\cdots,q_K^{n}\big)$, with 
\begin{equation}\label{MUT}
q_k^{n} = \frac{\mathcal{L}(\bm{\theta}_k^n|\mathcal{D}^n) }{\sum_{i=1}^K\mathcal{L}(\bm{\theta}_i^n|\mathcal{D}^n) }.
\end{equation}

The resampling criteria can be set to different conditions. For example, the resampling can be triggered if there has been $N^{\text{max}}$ iterations after the previous resampling step or if the likelihoods of more than half of the candidates are low.  As time goes by, the more data we have, the fitted model is expected to be more accurate and there is less risk of overfitting. 
In our work, we decrease the resampling frequency as we progress by resampling every $C\times 2^r$ iterations, where $C$ is a pre-specified constant and $r$ is the number of resamplings that we have executed. The detailed algorithm for the knowledge gradient policy with bootstrap aggregation is summarized in Algorithm \ref{KGBA}.

\begin{algorithm}
\caption{Knowledge Gradient with Bootstrap Aggregating}
\footnotesize
\SetKwInOut{Input}{input}\SetKwInOut{Output}{output}
 \Input{Initial set of parameters $\bm{\theta}_1,\cdots, \bm{\theta}_K$,  bids $p_1,\cdots, p_M \in [l,u]$, resampling base $C$.} 
 \textbf{initialize} $q_k^0 = 1/K$, $k=1,\cdots, K$, resampling counts $r=0$, $\mathcal{D}^0 = \emptyset$.\\
 \For{$n=0$ to $N-1$}{
 \If{$n = 2^rC$}{Get the bootstrapped models $\{\bm{\theta}_k\}_{k=1}^K$ from $\mathcal{D}^n$.\\
 Set the posterior distribution $q_k^{n} \leftarrow \frac{\mathcal{L}(\bm{\theta}_k|\mathcal{D}^n) }{\sum_{i=1}^K\mathcal{L}(\bm{\theta}_i|\mathcal{D}^n)}$  according to \eqref{MUT}.\\
 $r \leftarrow r+1.$
 }
 Receive the load $b^n$.\\
 Recommend a quote $p^n$ to the shipper and the carrier using the KG policy in \eqref{PKG}.\\
 Get responses $y^{n+1} = (y^{c,n+1}, y^{s,n+1})$ from the shipper and carrier, respectively.\\
 Update the posterior distribution $q_k^{n+1}$ according to \eqref{qq}:\\
 ~~~~~~~~~$q_k^{n+1} \propto q_k^n\sigma(y^{c,n+1}\cdot\bm{\alpha}_k^T\bm{x}^{c,n})\sigma(y^{s,n+1}\cdot\bm{\beta}_k^T\bm{x}^{s,n})$.\\
$\mathcal{D}^{n+1} = \mathcal{D}^{n} \cup \{b^n,p^n, y^{n+1}\}$.
}
\label{KGBA}
\end{algorithm}

\section{Model of the Fleet Management System with Advance Booking}\label{sec:sm}
This section begins by describes the base model which simulates the fleet and which details the load booking process, including the full booking profile which captures how far each shipper books loads in the future. We then present both the driver dispatch policy which draws on prior work using approximate dynamic programming, and the load acceptance policy, which uses a sophisticated stochastic lookahead model to estimate the probability that a load will actually be accepted.  Once the acceptance probability is estimated, we use a threshold policy where loads are accepted if the acceptance probability is over some minimum.  

For reasons of space, here, we just sketch the major elements of the base model which builds on \cite{SiDaGe09}, allowing us to focus on the load acceptance policy which is the element that is really new in this paper.

\subsection{The Base Model}

The base model, which is implemented as a simulator, can be written in the canonical form
\bn
\max_\pi \E \left\{\sum_{t=0}^T C(S_t,X^\pi(S_t))|S_0\right\}, \label{eq:baseobjectivemain}
\en
where 
\begin{align*}
S_t &=& \textwrap{The state of the system at time $t$, where $S_t$ consists of the state of drivers $R_t$, loads $L_t$, and the state of knowledge $K_t$ which captures our distribution of belief about how shippers and carriers might respond to prices in each traffic lane,}\\
S_0 &=& \textwrap{The initial state, which contains all deterministic information along with a Bayesian prior on the shipper and carrier response curves,}\\
x_t &=& \textwrap{$(x^D_t,x^L)$,}\\
x^D_t &=& \textwrap{Vector of decisions to assign drivers to loads,}\\
x^L_t &=& \textwrap{Vector of decisions to accept loads offered to the carrier,}\\
X^\pi(S_t) &=&  \textwrap{$(X^{\pi,D}(S_t), X^{\pi,L}(S_t))$,}\\
X^{\pi,D}(S_t) &=& \textwrap{Decision rule, or policy, for assigning drivers to loads,}\\
X^{\pi,L}(S_t) &=& \textwrap{Policy for accepting offered loads.}
\end{align*}

In previous sections, the model updates are for every load, indexed by $n$, while here the base model is modeled over time. $X^\pi(S_t)$ is the policy, or set of decision rules, that govern both load acceptance, the choice of which driver should cover each load, and as a byproduct, which loads are actually moved.  

The evolution of the state variable is described by the system model (also known as a transition function) which we write as
\bn
S_{t+1} = S^M(S_t,X^\pi(S_t),W_{t+1}), \label{eq:basetransitionmain}
\en
where
\bns
W_t &=& \textwrap{Vector of all exogenous information that first becomes known by time $t$ (that is, this is information that arrived between $t-1$ and $t$).}
\ens
The most important exogenous information in this project includes new loads being called in between $t-1$ and $t$, $\hat{L}_t$, as well as decisions made by shippers and carriers to accept a load at an offered price.  

Drivers and loads are modeled at a high level of detail.  Drivers are represented by an attribute vector $a$ that captures features such as current location, home domicile, driver type, and driving history (needed to enforce work rules).  We let
\begin{align*}
a &=& \textwrap{Vector of attributes, where $a\in\Acal$, describing a driver, which includes current (or inbound) location, home domicile, single/sleeper/owner operator, equipment type, nationality (to capture cross-border restrictions), target time to return home, driving hours, duty hours, and 8-day driving history,}\\
R_{ta} &=& \textwrap{Number of drivers with attribute $a$ at time $t$,}\\
R_t &=& \textwrap{$(R_{ta})_{a\in\Acal},$}\\
\Rhat_{ta} &=& \textwrap{Exogenous changes (additions or subtractions) of drivers with attribute $a$ that we learn about at time $t$,}\\
\Rhat_t &=&  \textwrap{$(\Rhat_{ta})_{a\in\Acal}.$}
\end{align*}
The random vector $\Rhat_t$ could be used to represent drivers entering and leaving the network (a process that we do not model in this paper), but it can also be used to model random travel times, where new information on the progress of a driver could be translated to subtracting a driver arriving at 10am.

Loads are described by the attribute vector $b$ which includes origin and destination, pickup and delivery windows, and other load characteristics.  We let
\begin{align*}
L_{tt'b} &=& \textwrap{$(L^O_{tt'b}, L^A_{tt'b})$}\\
       &=& \textwrap{Number of offered $(L^O_{tt'b})$ and accepted $(L^A_{tt'b})$ loads with attribute $b$ known at time $t$ that is available to be moved at time $t' \geq t$,}\\
L_{tt'} &=& \textwrap{$(L_{tt'b})_{b\in\Bcal}$},\\
\Lhat_{tt'b} &=& \textwrap{New (offered) loads becoming available with attribute $b$ between $t-1$ and $t$ to be picked up at time $t'$ (normally we would put $t'$ in the attribute vector $b$, but the lag between the call-in time $t$ and the earliest pickup time $t'$ plays a particularly important role in this paper),}\\
\Lhat_{tt'} &=& \textwrap{$(\Lhat_{tt'b})_{b\in\Bcal}$, a stochastic process of exogenous loads.}
\end{align*}
We have to design a load acceptance process to choose the loads from $\Lhat_{tt'}$ which we will commit to moving, which moves then into the vector $L_{tt'} = (L_{tt'b})_{b\in\Bcal}$.

This notation allows us to define our state variable as
\bns
S_t = (R_t,L_t, K_t),
\ens
where $K_t$ is our belief about the price curves for shippers and carriers introduced in Section \ref{resampling}. Specifically, our belief state includes both the vector of current candidate model  parameters $(\bm{\theta_{t1}},\cdots, \bm{\theta_{tK}})$ and the posterior probabilities $\bm{q}_t$, $K_t =(\bm{\theta_{t}},\bm{q}_t)$. It should be noted that Section 4 describes the model updates for each load, indexed by $n$, while here we are modeling a simulator steps forward in time increments so that the base model is indexed by $t$.

We are now ready to define the decision variables in more detail.  Let
\begin{align*}
\Dcal &=& \textwrap{The set of decisions that can be made at time $t$ = $\big(d^h, \Dcal^L)$,}\\
d^h &=& \textwrap{Decision to hold a driver,}\\
\Dcal^L &=& \textwrap{Set of decisions to move a load, where an element $d\in\Dcal^L$ corresponds to an attribute $b_d\in\Bcal$, which means that $d\in\Dcal^L$ refers to moving a load with attribute $b_d$,}\\
x^D_{tad} &=& \textwrap{The number of drivers with attribute $a$ are acted on with a decision $d$ at time $t$,}\\
x^D_t &=& \textwrap{$(x^D_{tad})_{a\in\Acal, d\in\Dcal},$}\\
x^L_{tt'b} &=& \textwrap{$1$ if we accept a load offered at time $t$ to be moved at time $t'$ with attribute vector $b$, $0$ if we decline the load.}
\end{align*}
The decisions are constrained by
\bn
\sum_{d\in\Dcal} x^D_{tad} &   =  & R_{ta},\label{eq:constraint1}\\
\sum_{a\in\Acal} x^D_{tad} & \leq & L_{ttb_d}, ~d\in\Dcal^L,\label{eq:constraint2}\\
x^D_{tad}                  & \geq & 0, \label{eq:constraint3}\\
x^L_{tt'b}                 & \leq & \Lhat_{tt'b}, \label{eq:constraint4}\\
x^L_{tt'b}                 & \geq & 0. \label{eq:constraint5}
\en
Constraint \eqref{eq:constraint1} requires that we act on all drivers with some decision (even if it is to do nothing), while constraint \eqref{eq:constraint2} limits the number of assignments to loads of type $b_d$ for $d\in\Dcal^L$ to the number of loads that have been accepted, represented by the vector $L_t$.  Constraint \eqref{eq:constraint4} limits the number of accepted loads by the number of loads that have actually been offered.  The policy $X^\pi(S_t)$ which determines $x^D_t$ and $x^L_t$ is required to satisfy these constraints.

The attribute vector $a$ can be quite complex (it may include up to 15 different elements).  We capture the dynamics of updating all the attributes of a driver using the attribute transition function which we write as
\bns
a' = a^M(a, d),
\ens
where $a'$ is the attribute of a driver with initial attribute vector $a$ after being assigned to a load with attribute vector $b$.  The attribute $a'$ is the post-decision attribute vector, which is the attribute vector we expect from assigning a driver with attribute $a$ to a load with attribute $b$, but without any new information.  We  define
\bns
\delta_{a'}(a,d) &=& \left\{\begin{array}{ll}
                            1, & \mbox{if $a^M(a,d) = a'$,} \\
                            0, & \mbox{otherwise.}
                            \end{array}  \right.
\ens
This allows us to write the post-decision resource vector as
\bn
R^x_{ta'} &=& \sum_{a \in \Acal} \sum_{d \in \Dcal} \delta_{a'}(a,d) x_{tad}.        \label{eq:transition1}
\en
We can then write our next pre-decision resource vector using
\bn
R_{t+1,a} = R^x_{ta} + \Rhat_{t+1,a}, ~a\in\Acal.    \label{eq:transition2}
\en
The transition function for loads is modeled using
\bn
L_{t+1,t'b} &=& L_{tt'b} + x^L_{tt'b}, ~t'> t. \label{eq:transition3}
\en
Note that eventually any load represented by $L_{tt'b}$ for $t'>t$ eventually becomes $L_{ttb}$, at which point it may be dispatched.  We assume that any loads in $L_{ttb}$ that need to be picked up at time $t$, but which are not, are lost but incur a real cost (not just a soft penalty) since the carrier must pay a broker to cover these loads.

Equations \eqref{eq:transition1}-\eqref{eq:transition3}, along with the updates to the belief state, represent the transition function $S^M(\cdot)$.  Our next challenge is to design our the driver dispatch policy $X^{\pi,D}(S_t)$ and the load acceptance policy $X^{\pi,L}(S_t)$.

\subsection{The Booking Process}
The booking process $\Lhat_{tt'b}$ is created from a series of probability distributions estimated directly from a historical dataset.  While a detailed description is not central to this paper, it helps to understand some of the characteristics of the booking process that were captured:
\begin{itemize}
  \item The prebook process - In our dataset, approximately 50 percent of loads were called in to be picked up four or more days into the future.  We captured this prebook distribution up to 14 days into the future.
  \item Seasonality - The model captures day of week, week of year, as well as a variety of holiday effects (e.g. the number of days before major holidays).
  \item Outbound distributions - We estimate the probability distribution of freight out of each region, by shipper and total.
  \item Lane distributions - We finally estimate the distribution of loads in each lane, in a way that ensures that the total freight out of a region for a shipper matches the aggregate distribution for that shipper.
\end{itemize}
Simulations of loads from these distributions have been found to closely match historical behaviors.

\subsection{The Driver Dispatch Policy}
The simplest policy for assigning drivers to loads is to use a simple myopic assignment problem, but this ignores the value of drivers in the future, which can depend on the attributes of the driver.  For example, some regions may have better loads for getting a driver home, or may have longer loads that are better suited if the truck is operated by a two-person sleeper team. 
To capture the impact of drivers on the future, we use the method based on approximate dynamic programming presented in \cite{SiDaGe09}. 

We define the objective function using 
\begin{align*}
c_{tad} &=& \textwrap{The contribution generated by applying decision $d$ to drivers with attribute vector $a$ at time $t$. This contribution captures empty movement costs, loaded revenues, and any service penalties from picking up or delivering loads early or late.}
\end{align*}

We then define the driver dispatch decision variable by $x_{tad}$ which is the number of drivers with attribute $a$ which we act on with a decision of type $d$.  There are two classes of decisions: hold (do nothing), or assign a driver to a load of type $b\in\Bcal$.  These decisions are determined by the driver dispatch policy which we denote by $X^{\pi,D}(S_t)$ which we determine using approximate dynamic programming by solving
\begin{equation}\label{adp}
X^{ADP,D}(S_t) = \argmax_{x_t \in \Xcal_t} \sum_{a\in\Acal} \sum_{d\in\Dcal} c_{tad}x_{tad} + \Vbar_t(R^x_t), 
\end{equation}
where
{\small
\begin{eqnarray}
\Vbar_t(R^x_t) &=& \sum_{a'\in\Acal} \vbar_{ta'} R^x_{ta'}, \nonumber \\
               &=& \sum_{a'\in\Acal} \vbar_{ta'} \left(\sum_{a \in \Acal} \sum_{d \in \Dcal} \delta_{a'}(a,d) x_{tad}\right),\label{adp2}
\end{eqnarray}}
and $\vbar_{ta'}$ is an estimate of the marginal value of a driver with attribute $a'$ at time $t$.  

Equations \eqref{adp} - \eqref{adp2} can be solved using a standard integer programming package.  Dual variables from this optimization problem are used to estimate the coefficients $\vbar_{ta'}$ (this is fully described in \cite{SiDaGe09}).  This policy was found to produce very high quality solutions that closely calibrated against the performance of a top truckload carrier.

A key feature of our model, not represented in \cite{SiDaGe09}, is that loads are called in at a time $t$ to be served at a time $t' \geq t$ that might be up to 14 days in the future (approximately half the loads are to be picked up four or more days in the future), while in the earlier work,  all loads were called in to be served right away. This means that we have to make commitments now to move a load in the future, even though we do not know the status of drivers in the future.   In this paper, we explicitly model the lags in the arrivals, represented by the variable $\Lhat_{tt'}$.  The variable $\Lhat_{0t'}$ represents all loads known initially over the horizon.  There is a lot of uncertainty in truckload trucking which complicates the process of making these advance commitments.

\subsection{The Load Acceptance Policy}
The more substantive difference between this paper and \cite{SiDaGe09} is that we introduce the dimension of explicitly deciding whether an offered load, which typically needs to be moved at some time in the future.  The challenge here is that whether we can accept a load in the future depends in part on whether we have a driver who might be able to handle the load.  Then there is also the issue of whether we still want to handle the load (other loads may be more profitable).

Truckload carriers typically use either simple rules (accept the load if the carrier has moved loads in that lane in the past), or calculations based on estimates of the net flow into the origin of the load (from which we can see if there may be drivers available) to net flow out of the destination (if we have too many loads going into a destination we may want to decline the load).  Such heuristics ignore the ability to move drivers into and out of neighboring regions, or the need to get drivers home.

We represent the load acceptance decisions  by $x^L_{tt'b}$ which is the number of loads in the offered set $\Lhat_{tt'b}$ which is added to the set of committed loads $L_{tt'b}$.  Acceptance decisions are made by the load acceptance policy that we designate $X^{\pi,L}(S_t)$.  Our policy uses an estimate of the probability that an offered load (which may be picked up in the future) will actually be moved.  

We have chosen to use a full stochastic lookahead policy to estimate the probability that a load will be accepted.  This is done by simulating our ADP policy $X^{ADP,D}(S_t)$ into the future, but this is done by performing 20 simulations of a {\it lookahead} model, which is a sampled version of the base model in equations \eqref{eq:baseobjectivemain}-\eqref{eq:basetransitionmain} which assumes that all offered loads are accepted.   Our lookahead model closely parallels the base model, with the following changes:

\begin{itemize}
\item To capture the parallel problems while recognizing that the lookahead model is a distinct model from the base model, all the variables in the lookahead model have tilde's and double time indices, as in $\xtilde_{tt'}$ for decisions or $\Stilde_{tt'}$ for state variables.
\item The first time index represents the time within the base model that the lookahead model is being formulated.  The second time index, $t'$, is the time within the lookahead model, which spans time periods $t'=t, \ldots, t+H$.  For doubly time-indexed variables in the base model such as $x^L_{tt'b}$ or $\Lhat_{tt'b}$, the corresponding variable in the lookahead model would be $\xtilde^L_{tt't''b}$ or ${\tilde \Lhat}_{tt't''b}$.
\item We let $\omegatilde_t\in\Omegatilde_t$ represent a sample of $\Wtilde_{tt}, \ldots, \Wtilde_{t,t+H}$, where $\Omegatilde_t$ is a set of 20 sample paths constructed from the load distributions that were estimated from history.  Just as there is a probability space $(\Omega,\Fcal,\Pcal)$ for the base model, there is a probability space $(\Omegatilde_t, \Fcaltilde_t, \Pcaltilde_t)$ for each lookahead model triggered at time $t$ with its own set of filtrations $\Fcaltilde_{tt'}\subseteq \Fcaltilde_{t,t'+1}$.  Also, just as every variable indexed by $t$ is $\Fcal_t$-measurable, every lookahead variable indexed by $tt'$, such as the state $\Stilde_{tt'}$, is $\Fcaltilde_{tt'}$-measurable, and all of these variables are $\Fcal_t$-measurable in the base model.
\item The lookahead model uses loads sampled from the load distributions estimated from history.  However, the driver dispatch and load acceptance policies are simulated in the base model using actual historical data.
\end{itemize}
The driver dispatch decisions are made using
\bns
\xtilde^D_{tt'} = X^{ADP,D}(\Stilde_{tt'}),
\ens
where the ADP policy $X^{ADP,D}(\cdot)$ is given by equations \eqref{adp}-\eqref{adp2}, using the decision variables $\xtilde_{tt'}$ and data of the lookahead model, but where the value function approximations $\Vbar_t(R^x_t)$ are from the base model.  However, the load acceptance policy in the lookahead model consists of a simple rule: accept all loads.  Thus, instead of using $x^L_{tt'b}$ in equation \eqref{eq:transition3}, we simply use all the offered loads and then observe which of these loads are accepted.

We need to compute the probability that a load with attribute $b$ is accepted in the lookahead model.  Let $\xtilde^D_{t,t'ad}(\omegatilde)$ be the decision made in the lookahead model (initiated at time $t$ in the base model) using our ADP policy while following sample path $\omegatilde\in\Omegatilde_t$.  Now compute a sampled estimate of the probability that loads with attribute $b_d$ for decisions $d\in\Dcal^L$ are covered in the lookahead model using
\bns
\rhohat_{t,t'b_d}(\omegatilde_t) = \frac{\sum_{a\in\Acal} \xtilde^D_{t,t'ad}(\omegatilde)}{L_{t,t'b_d}}, ~d\in\Dcal^L.
\ens
We run the lookahead model in parallel for each $\omegatilde_t \in\Omegatilde_t$, where the set $\Omegatilde_t$ would have 20 randomly sampled paths.  We then average across all the sample paths to obtain the estimated probability that an offered load will actually be moved using
\bns
\rhobar_{t,t'b} = \frac{1}{|\Omegatilde_t|} \sum_{\omegatilde_t \in \Omegatilde_t} \rhohat_{t,t'b}(\omegatilde_t).
\ens

We now propose a simple parametric policy for load acceptance of the form
\bn
X^{\pi,L}(S_t) &=& \left\{\begin{array}{ll}
                            1, &  \mbox{If $\rhobar_{t,t'b} \geq \theta^{accept}$} \\
                            0, & \mbox{Otherwise,}
                            \end{array}  \right.~~b\in\Bcal.
\en
We note that there is a lot of knowledge in the probability $\rhobar_{t,t'b}$ that loads of type $b$ to be moved at time $t'$ will be accepted.  Since we are running a full simulation using the same dispatch policy as the base model, we avoid all the standard approximations used by even the most sophisticated carriers.  This logic captures the availability of drivers, not just in the region near the origin of the load but across the entire network.  It also considers whether a load might help a driver get home, and also balances the profitability of the load against all other loads.

\section{Case Study}\label{sec:cs}
We test our methodology on the data from a major freight brokerage with the goal of assessing how well our bidding policy balanced ``learning while earning'' in an online environment.  


\subsection{Shipper and Carrier Acceptance Model}\label{Discrete}
We use discrete choice models used to estimate  the probability of acceptance as a function of price and other load attributes, carrier attributes and shipper attributes. Given a load $n$, $Y^{c, n} = 1$ denotes the acceptance by a carrier of the broker's quoted price, while $Y^{s, n} = 1$  means that the shipper has accepted the price. 

We begin by analyzing the data of truckload shipments managed by a U.S.-based freight brokerage company, which spans from  08-Aug-2015 to the end of 13-Sept-2015 and contains approximately 10113 loads. Each data point contains the following information: called-in time of the load, origin and destination of the load, load type, travel distance, and  the price accepted by the shippers and the carriers. The origins and destinations have been aggregated up to  3-digit zip-codes, resulting in a total number of 619 different origins, 720 different destinations, and 3,013 traffic lanes with positive traffic flows. Reflecting regional patterns in production and consumption, these lanes tend to be unbalanced, producing a range of market prices across the country.   For this reason, we use lane-specific attributes as context information in the discrete choice model to explicitly model the heterogeneity and to exploit commonalities between lanes in the trucking network.   Specifically, if an origin/destination has more than 15 observations, we build a corresponding indicator variable, resulting in 199 origin indicator variables $ \mathbb{I}^O_1, \cdots,  \mathbb{I}^O_{199}$ and 
228 destination indicator variables $ \mathbb{I}^D_1, \cdots,  \mathbb{I}^D_{228}$. An explanation of this approach is that if a traffic lane involves a popular origin and a seldom visited destination, the potentially accepted quote is roughly the average  price accepted among the lanes with the same origin, since there is not enough information about the destination.

In specifying behavioral models for the shippers and carriers, other than the above mentioned indicator variables, we choose variables that capture load and lane attributes. These attributes together make up the contextual information  $b$. The model is largely defined by how these attributes affect acceptance price, which is the only decision  variable. The variables used in the discrete choice models are provided in Table \ref{model}. It should be noted that the decision variable $p$ is understood as a per-mile rate.  
\begin{table}[htp!]
\centering
\footnotesize
\caption{Variable Description \label{model}} 
\begin{tabular}{p{4cm}p{11.5cm}}
\toprule
Variable	& Description \\ \hline
$p$ & Price per mile. \\
$\mathbb{I}^T_i$ & Indicator variables of the equipment type required by the load, $i = 1,2,\cdots,5$, corresponding to ``Dry Van'', ``Flat Bed'', ``Refrigerated'', ``RGN'' and ``Step Deck''. Only one out of the five is one and others are 0. \\
Miles	& The total distance of the load.\\
MinDist	& An indicator variable describing the distance of the load: MinDist = 1 if $\text{Miles}\le 300\text{mi}$.\\
DailyLoad	& The averaged daily load of this traffic lane from historical data. \\
$\mathbb{I}^O_i$ &   Indicator variables of 199 different origins, $i = 1,2,\cdots,199$. \\
$\mathbb{I}^D_i$ &   Indicator variables of 228 different destinations, $i = 1,2,\cdots,228$. \\
DestDailyDemand & The averaged daily demand out bounded from the destination agregated from historical data.\\
\bottomrule
\end{tabular}
\end{table}

The full model calibration for shippers $f^s(b,p;\bm{\beta}) = \sigma\big(U(x^s(b,p)|\bm{\beta})\big)$ and carriers $f^c(b,p;\bm{\alpha})=\sigma\big(U(x^c(b,p)|\bm{\alpha})\big)$ is described as follows:
\begin{align*}
U(\bm{x}^c|\bm{\alpha}) &= \alpha_0 + \alpha_1\text{DailyLoad} + \alpha_2\text{DestDailyDemand}  + \alpha_3\text{MinDist}+(\alpha_4 \mathbb{I}^O_1 + \cdots + \alpha_{202} \mathbb{I}^O_{199})\\
&\qquad+ (\alpha_{203} \mathbb{I}^D_1 + \cdots + \alpha_{430} \mathbb{I}^D_{228})+(\alpha_{431}\mathbb{I}^T_1+\alpha_{432}\mathbb{I}^T_2 
+\alpha_{433}\mathbb{I}^T_3+\alpha_{434}\mathbb{I}^T_4+\alpha_{435}\mathbb{I}^T_5)\\
&\qquad+(\alpha_{436}p\cdot\mathbb{I}^T_1+\alpha_{437}p\cdot\mathbb{I}^T_2 
+\alpha_{438}p\cdot\mathbb{I}^T_3+\alpha_{439}p\cdot\mathbb{I}^T_4+\alpha_{440}p\cdot\mathbb{I}^T_5)\\
&\qquad+\alpha_{441}p\cdot\text{Miles}\cdot\text{MinDist} + \alpha_{442}p\cdot\text{DailyLoad}.\\
U(\bm{x}^s|\bm{\beta}) &=  \beta_0 + \beta_1\text{MinDist}+(\beta_2 \mathbb{I}^O_1 + \cdots + \beta_{200} \mathbb{I}^O_{199})+ (\beta_{201} \mathbb{I}^D_1 + \cdots + \beta_{428} \mathbb{I}^D_{228})\\
&\qquad+(\beta_{429}p\cdot\mathbb{I}^T_1+\beta_{430}p\cdot\mathbb{I}^T_2 
+\beta_{431}p\cdot\mathbb{I}^T_3+\beta_{432}p\cdot\mathbb{I}^T_4+\beta_{433}p\cdot\mathbb{I}^T_5)+\beta_{434}p\cdot\text{Miles}\cdot\text{MinDist}.
\end{align*}
 
 The carrier's model is richer since the cost of the load is highly uncertain due to follow-on loads associated with deadheading (moving an empty truck to the origin of the next load),  dwell time, as well as  the overall spatial and temporal variability of load demands. For example, the willingness of a carrier to haul from region A to region B will be higher if the truck has a high chance of getting an outbound load from B.  Second, in the data we have, in order to avoid overfitting, we use $l_1$ regularized logistic regression. When the regularization parameter is set to one, we  do not find commodity  types to be statistically significant as intercept variables in the shipper demand model, so we omit them in the final model specification. 
 
 \subsection{Simulated Experiments}\label{simulate}
\begin{figure}[htp!]
\begin{center}
  \subfigure[Carrier acceptance model]{\includegraphics[width=0.37\textwidth]{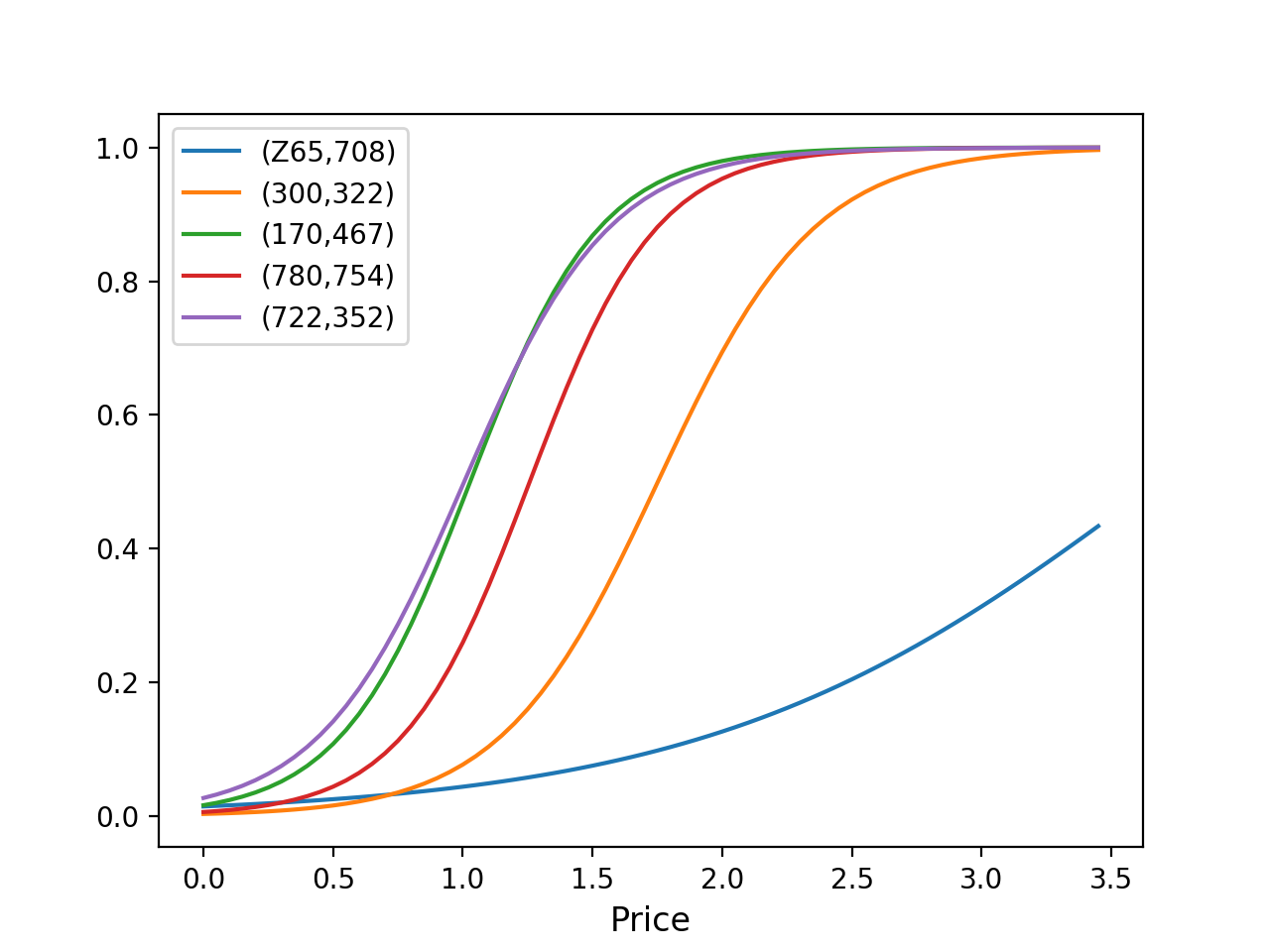}}~
  \subfigure[Shipper acceptance model]{\includegraphics[width=0.37\textwidth]{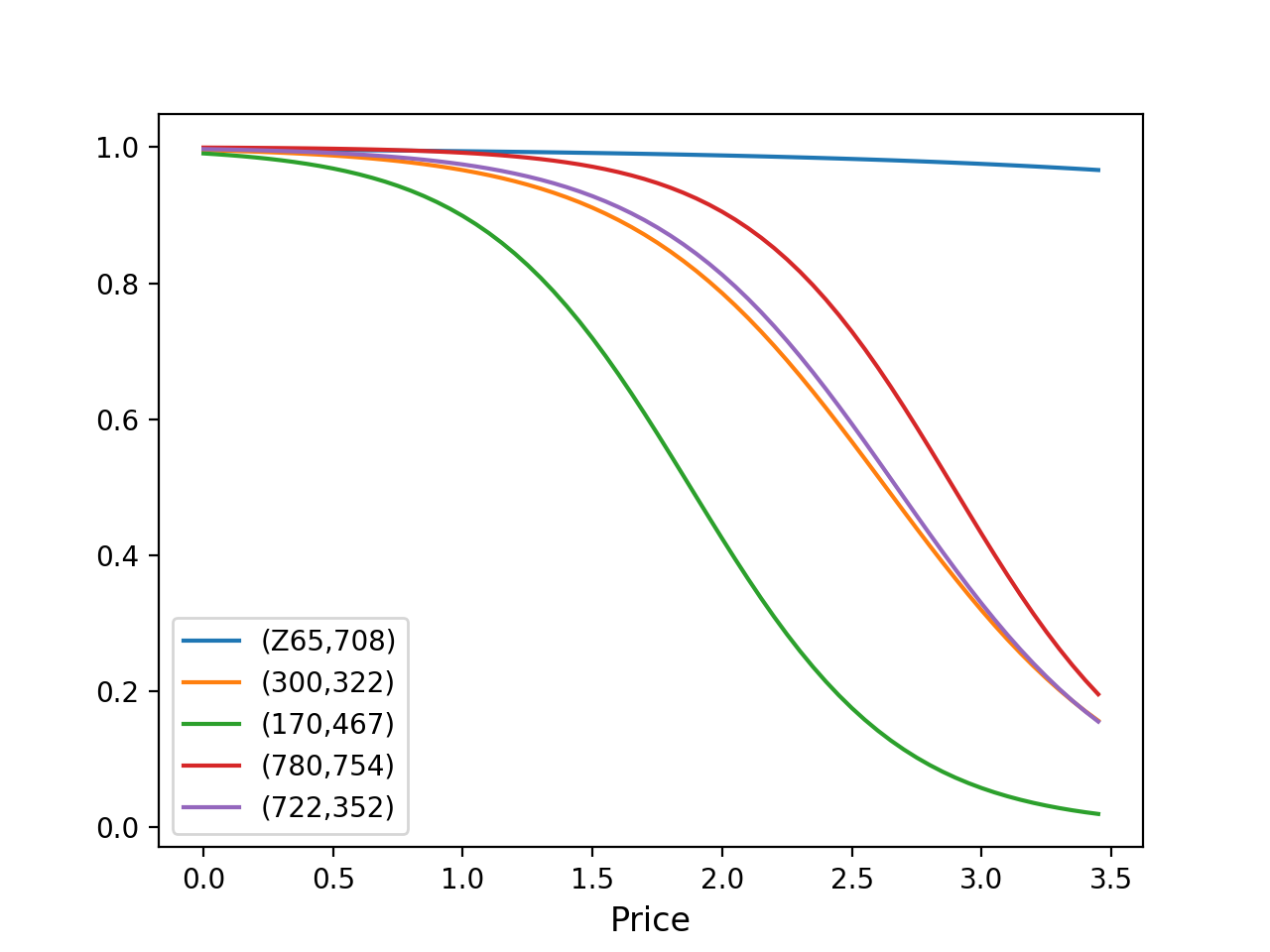}} 
    \caption{Shipper and carrier acceptance model for 5 different pairs of origins and destinations.\label{tl}}
 \end{center}
\end{figure}

 Evaluating an exploration/exploitation policy is difficult since we do not know the outcome of the bids that were not chosen for a particular load in the historical data \citep{chapelle2011empirical}. As a common practice in bandit literature, using real world context and freight shipment features in the truckload  dataset, we instead simulate the true outcomes for both shippers and carriers using the discrete choice models and a weight vector $\bm{\theta}^*$. This weight vector could be chosen arbitrarily, but it was in fact a perturbed version of some weight vector learned from real data.  Figure \ref{tl} illustrates the shipper and carrier acceptance probability as a function of the per-mile bid $p$, under the fitted parameter $\bm{\theta}^*$ by training the model using $l_1$ regularized logistic regression on the historical dataset. The figure was prepared using the corresponding acceptance models for carriers and shippers on five randomly chosen loads moving between different region pairs.  Although volumes across traffic lanes range from very low to very high, we can use the lane information in our acceptance models to explicitly capture the different behaviors across all the traffic lanes and exploit commonalities in the trucking  network.

\subsubsection{Comparison with Different Policies}
We sort the loads chronologically. For each load, we collect its attributes $b$, construct the explanatory variables as explained in Section \ref{Discrete}, and then use a policy to choose a bid. We then observe the responses of  the shipper and the carrier. The goal is to maximize the expected total revenue. We assume that prices fall in the range of (0,4] dollars per mile. In the experiments, we discretized the prices into 80 alternatives with $0.05$ as the increment. For each load, if we knew the truth parameter value $\bm{\theta}^*$, we can get the best mapping from any context $b$ to the optimal price by $\max_{p}pf(b, p;\bm{\theta}^*)$. This optimal value that can only be obtained in hindsight is used as the {\it optimal} benchmark such that we compare different policies against this maximum revenue achieved if we knew $\bm{\theta}^*$.  In other words, the comparison metrics is the expected cumulative regret achieved by different policies $\pi$ as time goes by, 
\begin{equation}
\mathcal{R}^{\pi}(n) = \sum_{t=0}^{n-1}\max_{p}pf\big(b^t, p;\bm{\theta}^*\big) -  \sum_{t=0}^n\pi(b^t)f\big(b^t, \pi(b^t);\bm{\theta}^*\big). 
\end{equation}

In the experiment, we set the number of candidate functions to be $K = 5$ and the resampling base as $C = 300$ indicating that the first resampling happens at the 300th iteration. We compare our knowledge gradient policy with pure exploitation (which recommends the quote that seems to be the best), and a static policy (Mean-Price) that always chooses the bid that is the mean of the accepted prices in the history.  The pure exploitation policy uses the same sampled belief model and the same bootstrapping procedures. The only difference with the KG policy is how it selects the bid at each time step. We also consider  a baseline policy that is termed the estimate-and-optimize  policy (Est-Opt). It uses the same logistic regression model describe in Section \ref{Discrete} as the prediction model. It starts with a specific set of parameter values which can be possible trained over past data. At each time step, it recommends the optimal price based on the estimated parameter values for the current load. While new information accrues at each time step, the parameter values are not updated until the next refresh. After a certain number of iterations, we will re-train the regression model taking into account the  newly observed data points. This is the conventional approach that is used in various industries \citep{phillips2005pricing, talluri2006theory} with constant refresh.
\begin{figure}[htp!]
\centering
\includegraphics[width=0.5\textwidth]{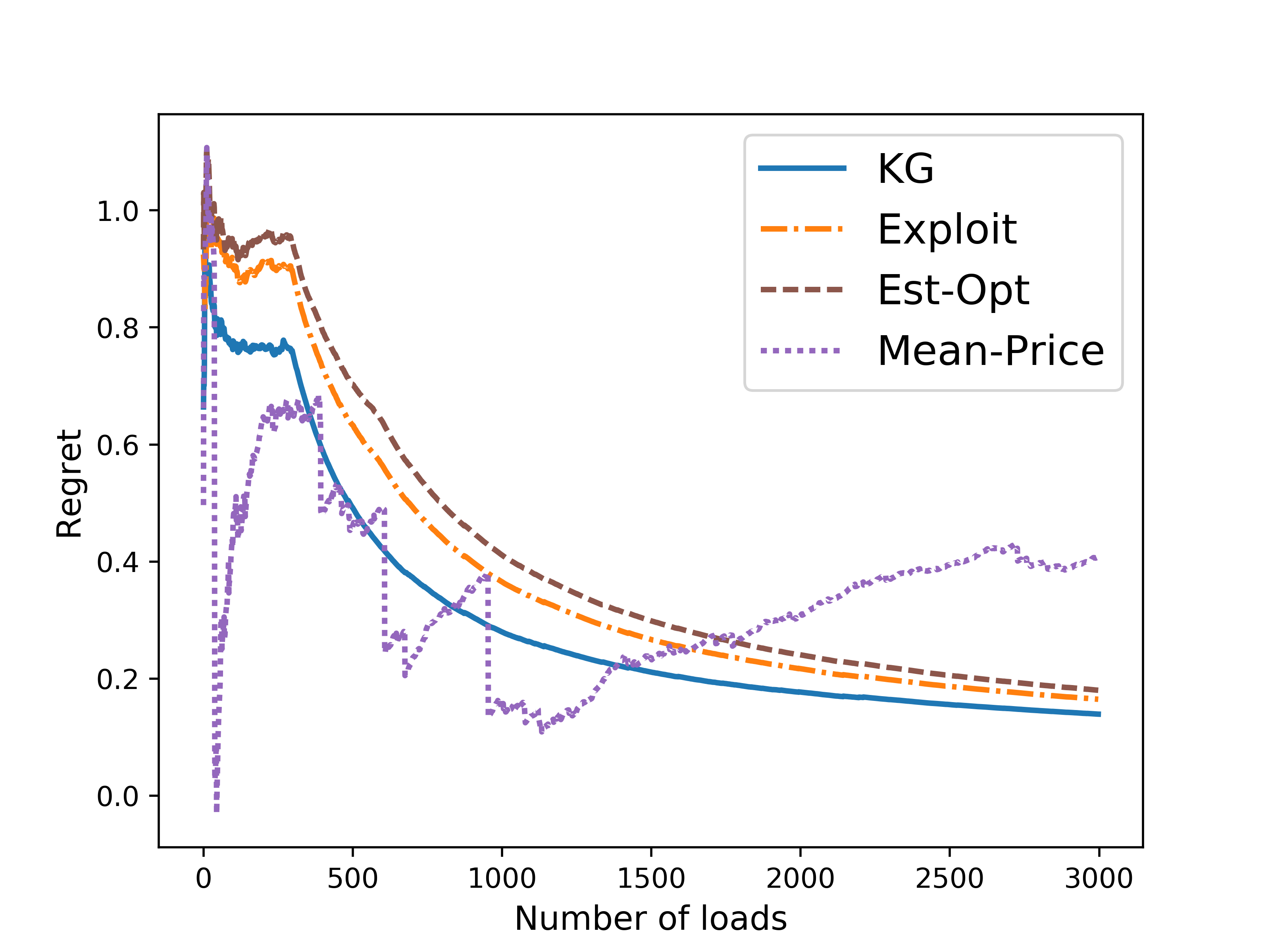}
    \caption{Comparison of different policies. \label{f:regret}}
\end{figure}

The experimental results are reported on 20 repetitions of each algorithm. We report the averaged cumulative regret $\mathcal{R}^{\pi}(n)/n$  in Figure \ref{f:regret} on the first 3000 iterations.  We can see that the KG policy yields the best performance. The Mean-Price policy has a constant performance since it does not learn anything on the way.  It emphasizes  the need for price experimentation to gather more information on other prices. It is worth mentioning that while other policies use the discretized prices from 0 to 4, the  Mean-Price policy uses the actual prices in historical data, which does not necessarily fall in this price interval. It includes lanes with prices of \$4.56 and \$7.13, although these are exceptional. The Est-Opt policy has a reasonable performance. This seems at first a bit odd since  no learning happens between two re-fitting steps and this myopic policy does not make any attempt to perform price experimentation to learn the response parameters. A possible explanation is that the change in context induces some level of exploration.  We also notice obvious breaking points at the iterations (e.g. 300 and 600) of resampling and/or refreshing in KG, Exploit and Est-Opt. This confirms our claim that in high-dimensional settings, it is highly unlikely that one of the pre-defined candidate functions is the true model.  The bagging procedure provides the possibility  to find more promising parameters and in return benefit for immediate earnings.  

The real value of the knowledge gradient is its rapid learning. We can see this by taking a close look at the intervals between two resamplings. The reduction in the regret of the KG policy is much sharper than other policies, indicating its ability to learn quickly. This is an important  advantage when the experiments are very expensive. The better performance of pure exploitation than Est-Opt also illustrates the benefit of a truly sequential model or an online learning model  in its rapid responses to the newly acquired information. 

\subsubsection{Effectiveness of Bagging.}
\begin{figure}[htp!]
\centering
\begin{tabular}{ll}
  \includegraphics[width=0.34\textwidth]{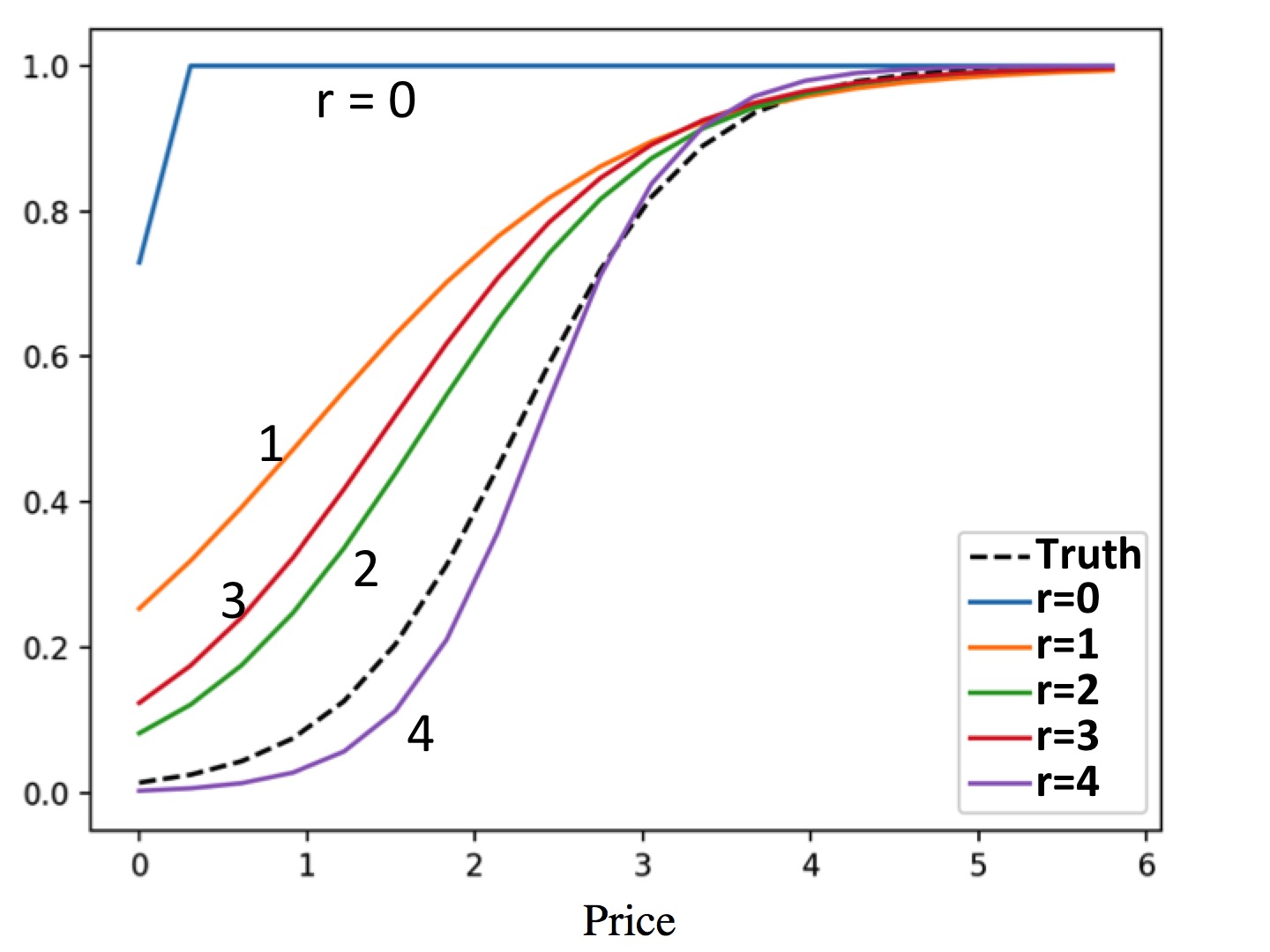} &
  ~~\includegraphics[width=0.34\textwidth]{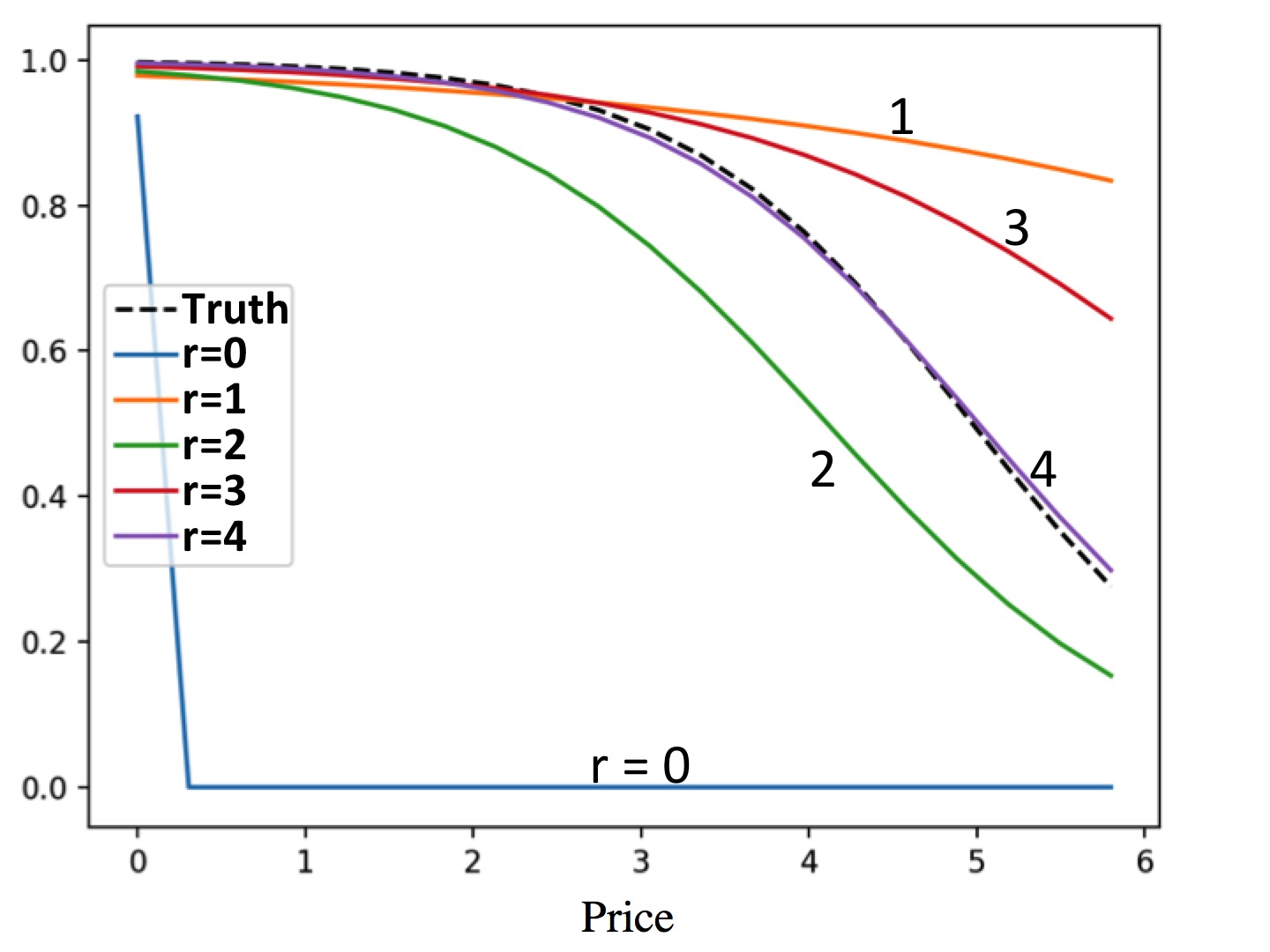} \\
    \includegraphics[width=0.34\textwidth]{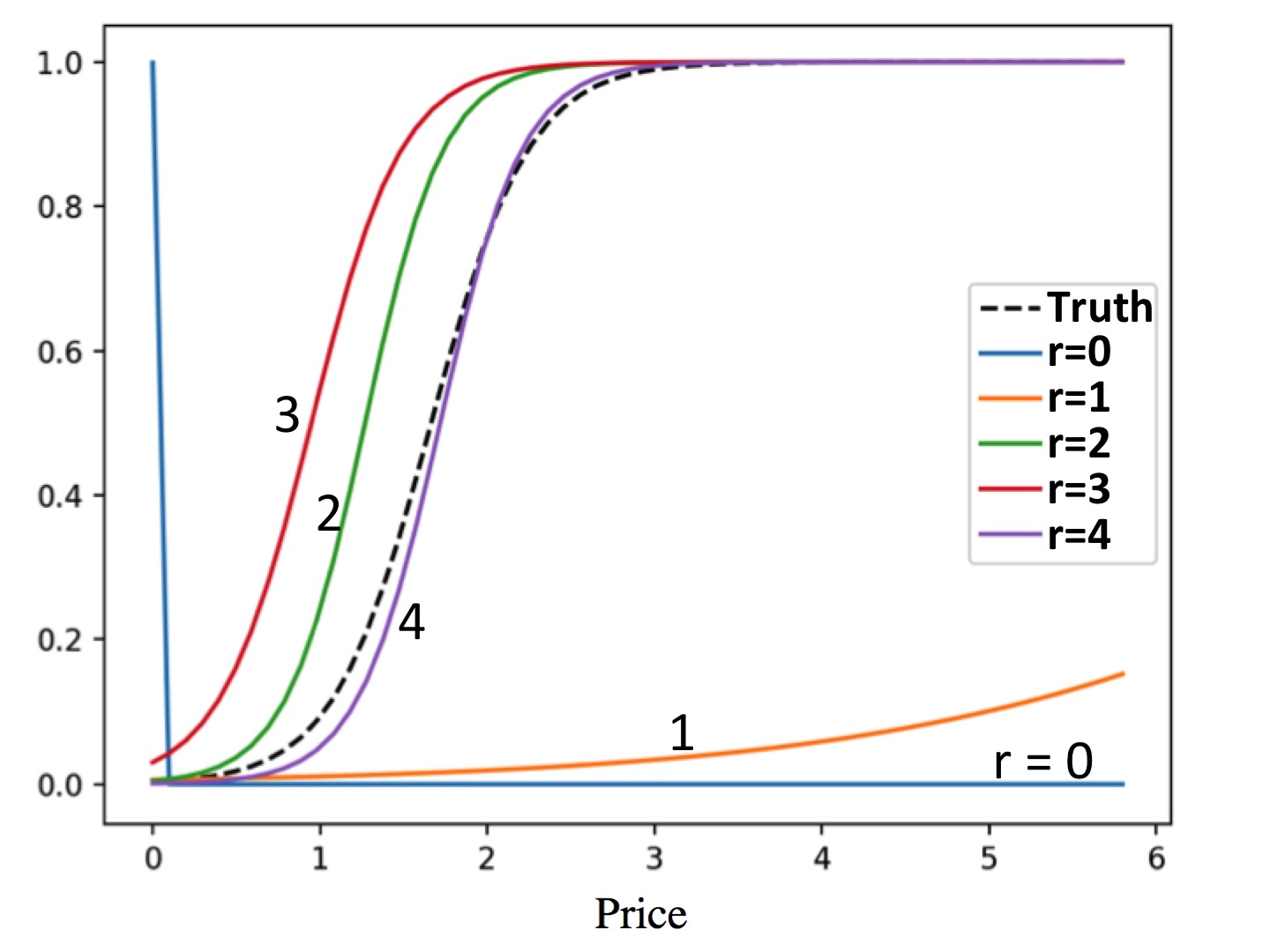} &
  \includegraphics[width=0.34\textwidth]{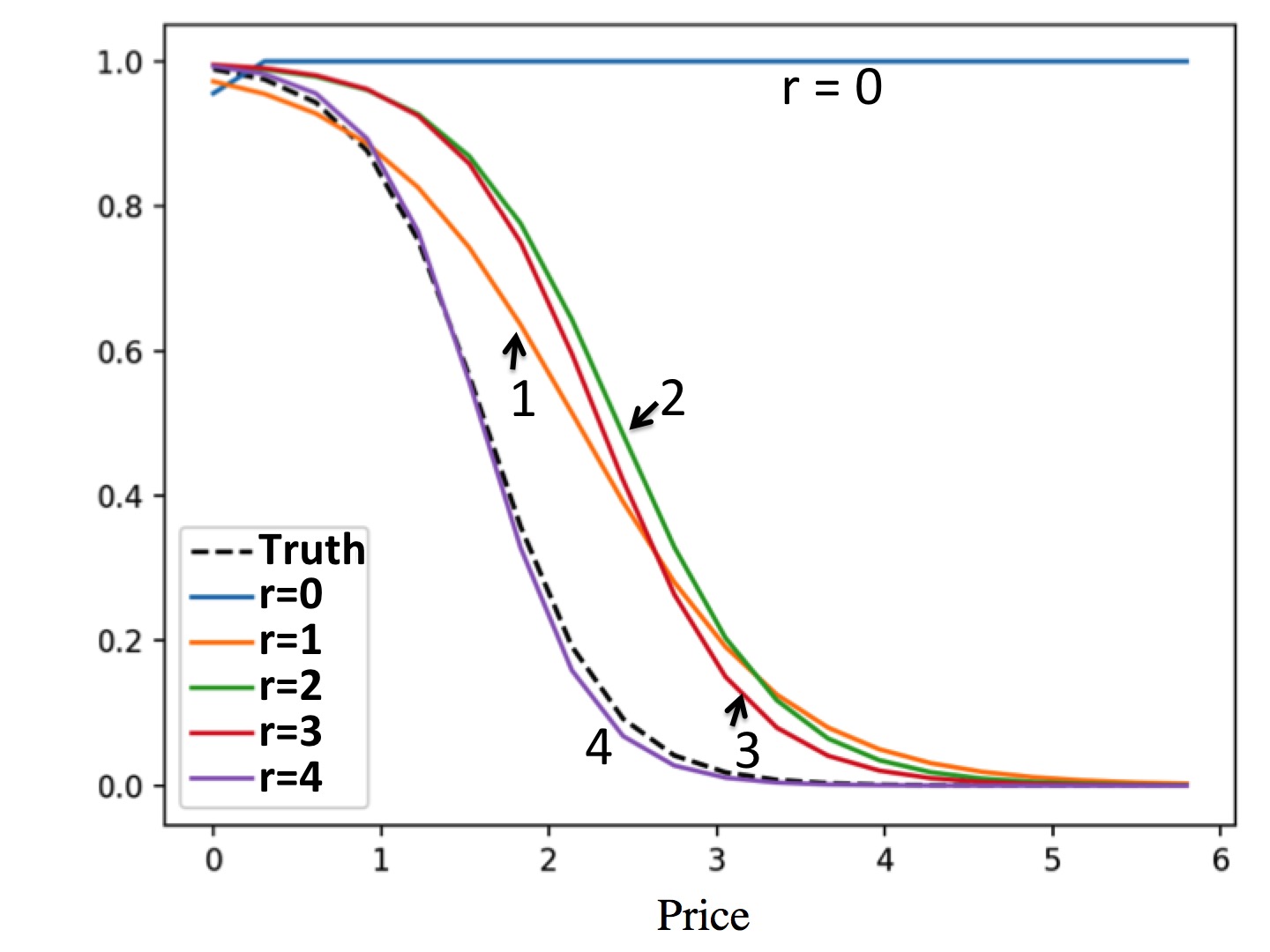} 
    \end{tabular}
    \caption{ Bid response prediction with the most probable model before each resampling step of two region pairs.  The black dashed line is the true model, and $r$ represents the number of previous resamplings. 
    \label{f:bagging}}
\end{figure}

As seen in Figure \ref{f:regret}, the breaking points of the regret curves at the resampling steps provide evidence of the effectiveness of bagging in finding the new sets of potential parameter values which promote locating the best acceptable bids. In Figure \ref{f:bagging}, we randomly select two loads from the dataset with different origins and destinations  and examine the fitting results  on the shipper and carrier acceptance model achieved by bootstrapping aggregation. After each bagging step, $K$ new parameter values are generated and the posterior distribution on the new set of candidate functions will continue to be updated as the knowledge gradient policy acquires new observation at each time step.  

We start with $K = 5$ randomly generated candidate functions. Before each bagging step, based on the posterior distribution, we find the most probable model $\hat{\bm{\theta}}$ in our current candidate set and plot the shipper and carrier acceptance probability of the selected loads, as a function of $p$.  Each line depicted in figure  \ref{f:bagging} is the fitted  bid response curve obtained before each bagging step $r$, with $r$ indicating the number of previous resamplings.  For example, $r = 0$ is the most probable model under the initial randomly selected candidate functions, right before the first resampling procedure occurs. The black dashed line is the true model under the simulated truth $\bm{\theta}^*$ that is unknown to the learner. We can see that  even though the initial estimates can be anywhere in the parameter space which makes it hopeless to recommend a good bid, by resampling parameters, we can continue to refine our estimation of the behavior of the shipper and the carrier.   We draw the fitting curves for other loads as well and observe similar results. This indicates that by using  a contextual model, we can optimize over the entire truckload network at the same time. 

This result together with Figure \ref{f:regret} provides an illustrative picture to demonstrate the behavior of the knowledge gradient policy in achieving the dual objectives of maximizing profits while learning the bid response curves.

\subsection{Experiments with Advance Booking Simulator}\label{real}
As recognized within the reinforcement learning community, evaluating the performance of bandit policies is inherently challenging since we do not observe the counterfactuals.  This makes it hard for researchers to reliably compare the performance of different algorithms. An ideal way to evaluate a bandit algorithm is to conduct field experiments, in which we actually run the bidding algorithm in reality and observe the responses from carriers and shippers. Section \ref{simulate} provides an evaluation method which is the common practice in bandit community, but by assuming the belief model is true, the issue of mis-specification is neglected, resulting in a possibly over-optimistic evaluation result.

In practice, carriers face complex factors that affect  driver management, including the availability  of a backhaul moves, hours of service commitments, the highly stochastic requests of loads,  the commitments to loads offered in the future, and the service penalty if an accepted load is not moved on time.  To provide a more realistic evaluation that can better mimic field experiments, Section \ref{sec:sm} builds on the driver management model presented in \cite{SiDaGe09} to produce an accurate simulation of a large-scale fleet that captures real-world operations at a very high level of detail.  In what follows, we conduct online evaluations of our proposed methodology using the simulator to provide  carrier acceptance/reject decisions. 

For every six hours spanning two weeks, the simulator offers a batch of loads to be moved to the broker. As explained in Section \ref{sec:sm}, the stochastic process of loads being called in  is an exogenous information process that is fitted from historical data to ensure that the pre-book distribution matches the historical distributions at different levels of aggregation.     The broker 
uses the KG policy to recommend bids for each non-committed load. The simulator calls the carrier's load acceptance policy described in Section \ref{sec:sm} and  determines whether the proposed bid is accepted or rejected by the carrier. In the meantime, we simulate the behavior of the shipper using the discrete choice model in Section \ref{Discrete} with a  true parameter value $\bm{\beta}^*$ and generate the shipper acceptance/reject decision to each bid.  The loads are strictly time-ordered since the load acceptance policy needs to take into consideration, for example, the number of loads accepted to be moved in the future and the number of drivers at each time step. The broker will use the carrier's and shipper's feedback to update his belief on the carrier and shipper response curve. The same procedure will be conducted for the next batch of loads called in within the next six-hour time period. 

We compare the performance of different policies in the bidding effectiveness. The baseline policy is the Est-Opt policy. We cannot use the Mean-Price policy since  there are many loads that have not been encountered in history. Other than the Exploit policy introduced earlier, we also extend two more state-of-the-art policies in the multi-armed bandit literature to our sampled belief model with bootstrapping aggregation, Thompson sampling \citep{thompson1933likelihood}, and its enhanced version, Optimistic Thompson sampling \citep{may2012optimistic}. Thompson sampling (TS) randomly
selects an action (bid) according to the current distribution of belief about its true value. In the case of a sampled belief model, at each time step, we can achieve this by randomly drawing a candidate function from the posterior multinomial distribution and select the price that is the best under the sampled function. Optimistic Thompson sampling (Opt-TS), modifies the score of each price to the maximum of the sampled value and its mean value under the posterior distribution \citep{may2012optimistic}. For Exploit, KG, TS and Opt-TS, the number of candidate functions is set to $K=5$ and the resampling base is set to $C = 100$. 
 \begin{figure}[htp!]
\centering
\includegraphics[width=0.6\textwidth]{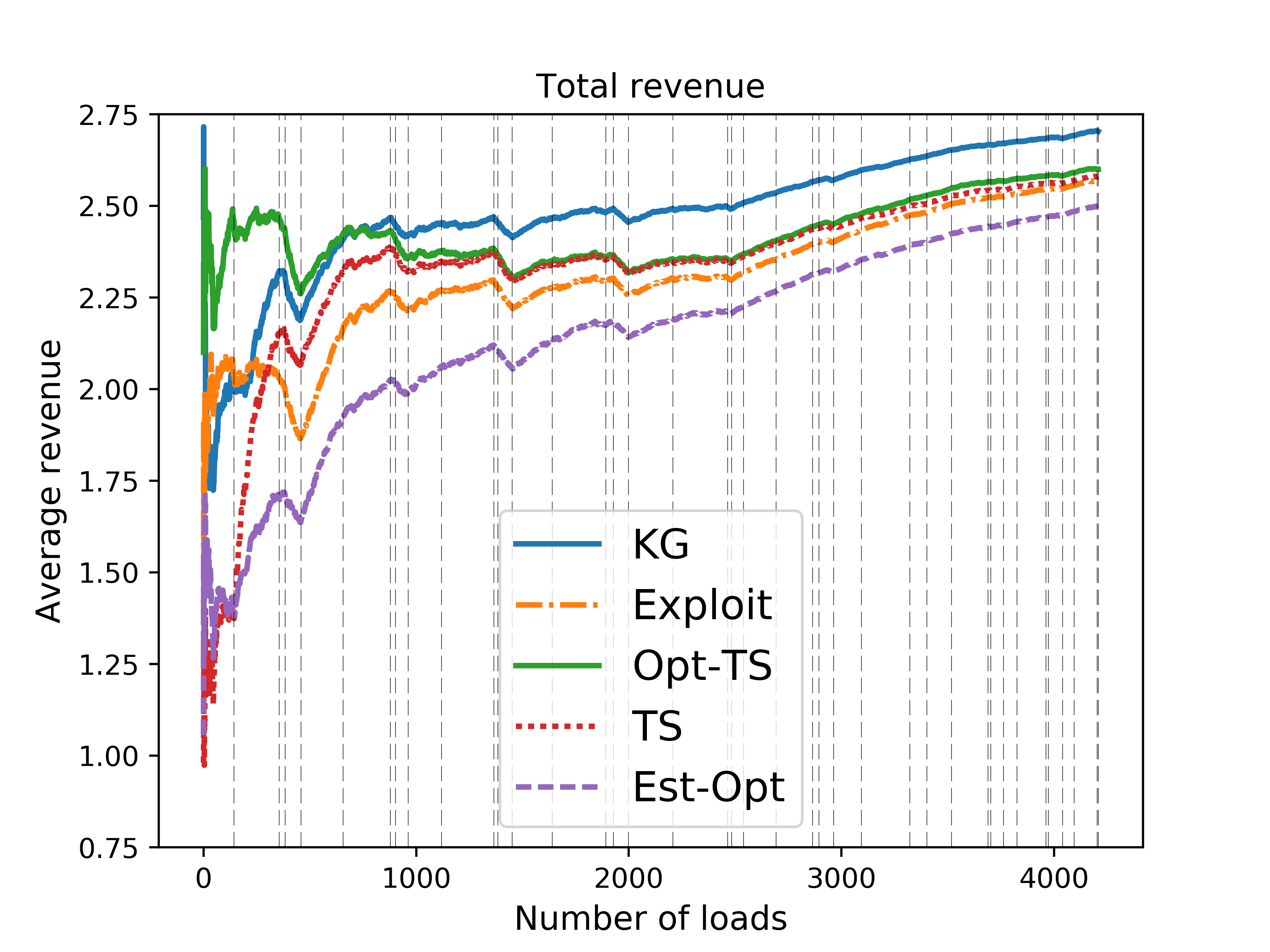}
    \caption{Comparison of different algorithms on total revenue using advance booking simulator.\label{real}}
\end{figure}

 Figure \ref{real}(a) reports the average revenue $\mathcal{G}^{\pi}(n)/n$ in order of time, where
\begin{equation*}
\mathcal{G}^{\pi}(n) = \sum_{t=0}^{n-1}P^\pi(b^t)\cdot Y^c\big(b^t, P^\pi(b^t)\big)\cdot Y^s\big(b^t, P^\pi(b^t)\big). 
\end{equation*} We repeat each policy on 10 different sample paths and report its average performance.  The x-axis is the sequence of 4209 upcoming loads spanning 2 weeks. The vertical dotted gray lines indicates the end of each 6-hour batch. For example, the first 144 loads depicted in  Figure  \ref{real} are the loads  called in during the first 6-hour period.  We can see from the figure that KG has the highest performance in the total revenue gained.    The better performance of pure exploitation than Est-Opt confirms our finding  that since pure exploitation has the ability to react to newly acquired information, it can use the new information to make better decisions. Not surprisingly, TS yields satisfactory performance, which has been empirically verified in other work \citep{chapelle2011empirical,agrawal2012thompson}, although in different application contexts and with different belief models.
Optimistic Thompson sampling achieves a slightly better performance than Thompson sampling, which illustrates the payoff gains using this optimistic version.  


We now turn to studying the cumulative number of bids that were accepted by both shipper and carrier. It is reasonable to make the assumption that within the 2-week time period, the stochastic process of loads calling in does not depend on the broker's bidding strategy or the broker's reputation.  However, the reality is that the freight industry depends heavily on relationships and reputation. If most of the loads called in are not fulfilled, it is hopeless to establish a long-term relationships.  Thus, the broker needs to maintain a steady rate of contract acceptance. 

\begin{figure}[htp!]
\centering
  \subfigure{\includegraphics[width=0.328\textwidth]{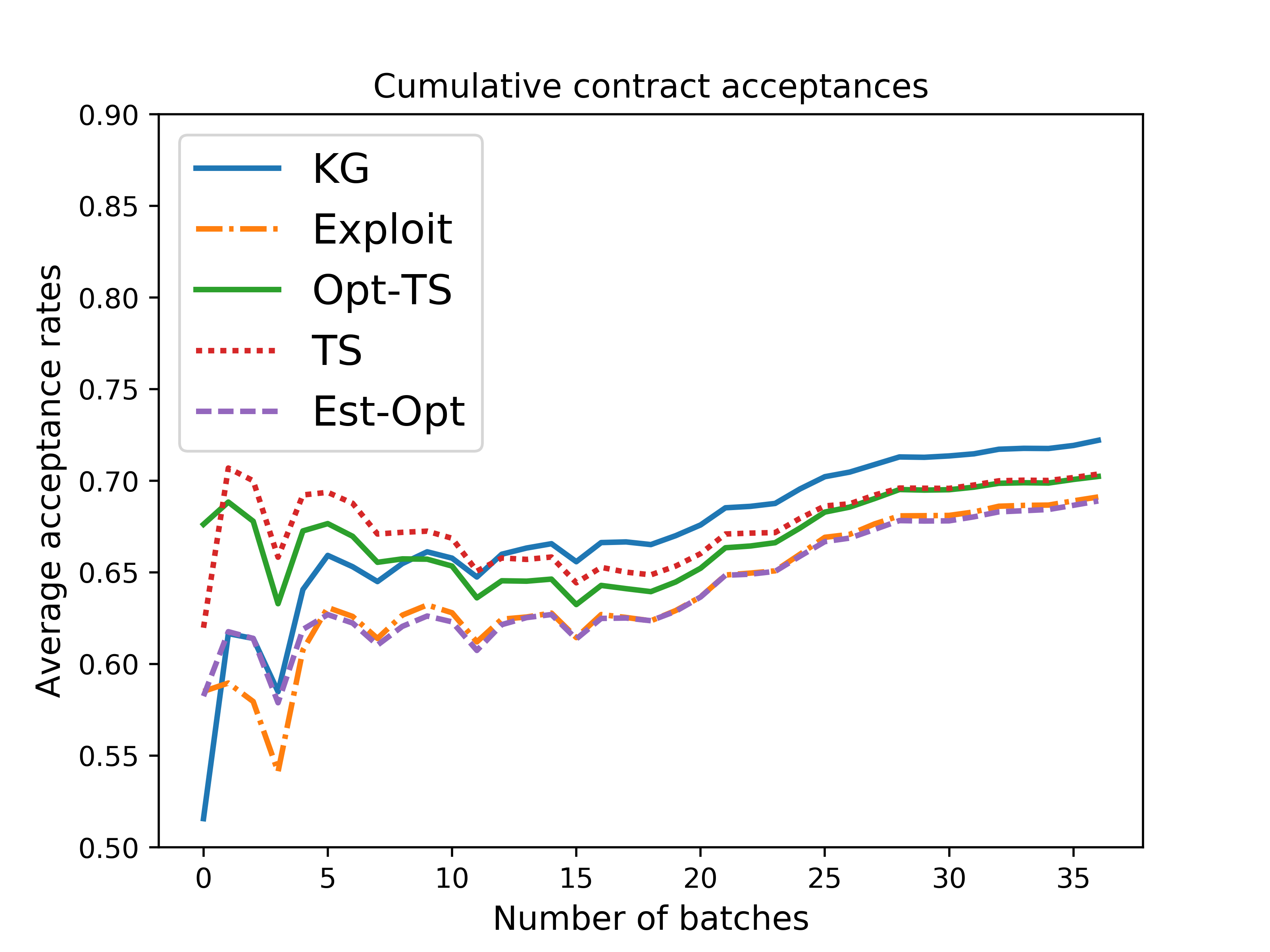}} 
    \subfigure{\includegraphics[width=0.328\textwidth]{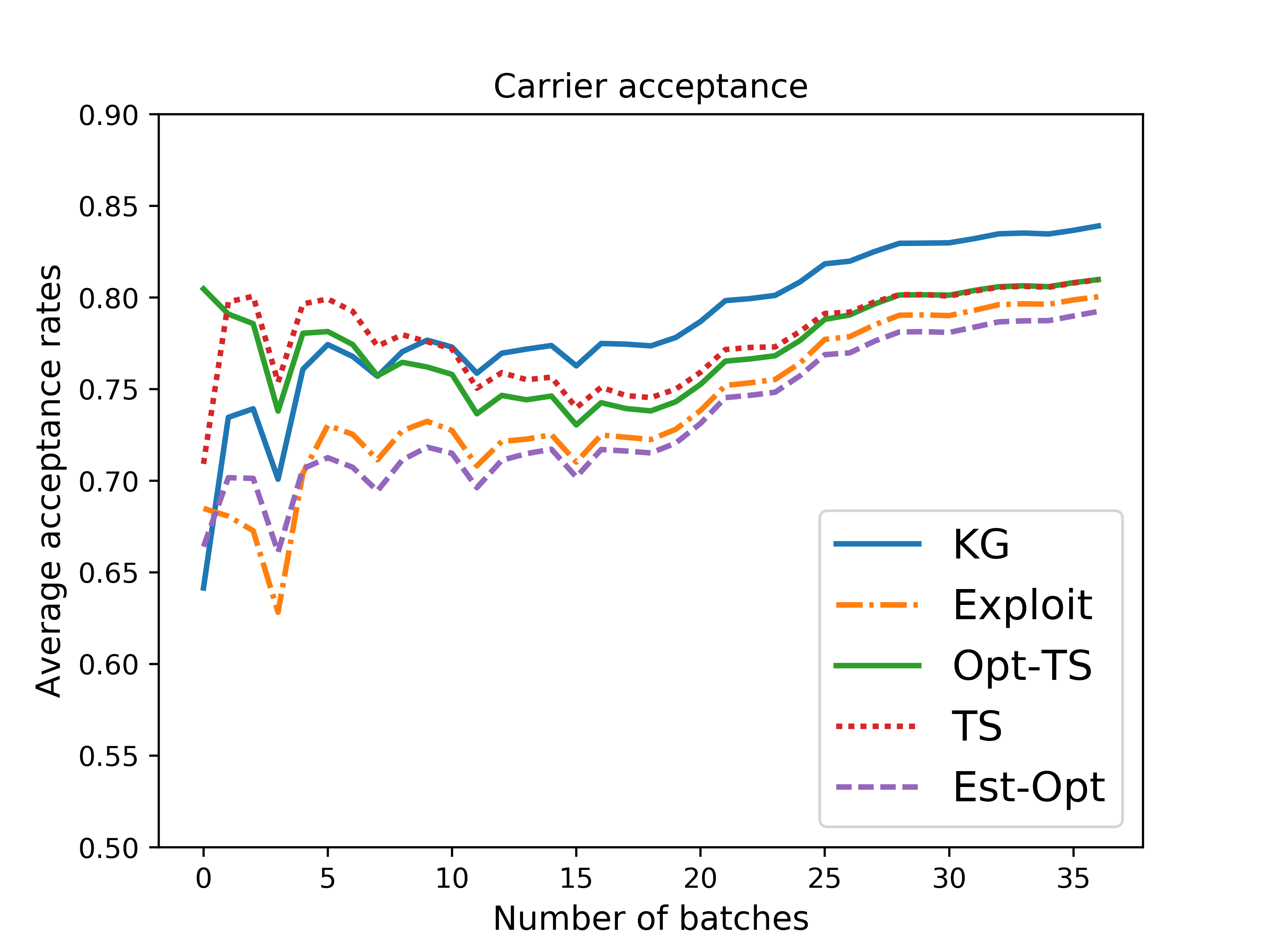}}
  \subfigure{\includegraphics[width=0.328\textwidth]{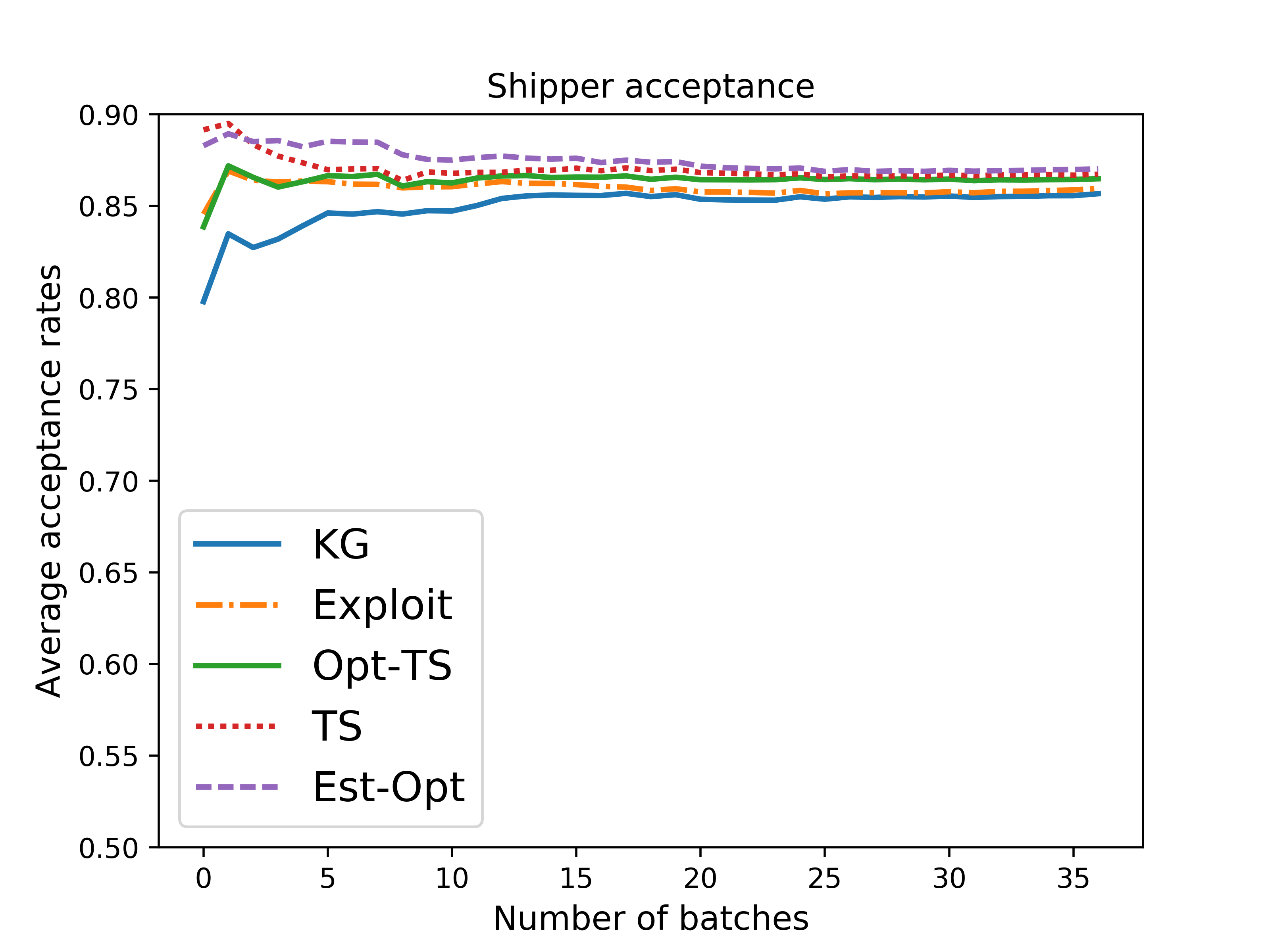}} 
    \caption{Acceptance rate achieved by different policies. Left: acceptance by both the carrier and the shipper; Middle: carrier acceptance;
Right: shipper acceptance.\label{real_prob}}
\end{figure}

To better understand the bidding efficiency, in Figure \ref{real_prob}, we report average acceptance by both, along with acceptances by shippers and carriers separately. The acceptance rate   $\mathcal{P}^{\pi}(n)/n$ is reported  at the end of each batch, with $$\mathcal{P}^{\pi}(n) = \sum_{t=0}^{n-1} Y^c\big(b^t, P^\pi(b^t)\big)\cdot Y^s\big(b^t, P^\pi(b^t)\big). $$  
Other than the largest total revenue achieved as illustrated in Figure \ref{real}, the KG policy yields the highest contract acceptance rate,  with a 3\% margin above the second best TS policy.  We can see from Figure \ref{real_prob} that the shipper acceptance curve is smoother while the carrier acceptance rates exhibit choppy behaviors among all the policies. One explanation is that shipper behavior is simulated using a logistic curve, while carrier behavior is determined by the dispatch logic using approximate dynamic programming, which means that the logistic curve is just an approximation of the carrier response.   The carrier acceptance across all the policies is lower in the middle of the two weeks, reflecting the need for the carrier to dispatch previously accepted loads. 


\section{Conclusion}
This paper presents, for the first time, a formal model of the dynamic bidding process for freight brokerages for truckload trucking, where we handle both the problem of advance booking and pricing.  We develop a contextual knowledge gradient policy to quickly learn the bid response curves for carriers (buyers) and shippers (sellers).  The knowledge gradient policy requires estimating a high-dimensional, spatially distributed belief model to capture the response to bids as a function of origin, destination and other attributes of the loads.  This required developing a novel resampling algorithm for efficiently computing the imbedded expectation of the maximum of an expectation of a nonlinear response function.  We show that the knowledge gradient policy produces asymptotically unbiased estimates of the shipper and carrier response curves.

The learning method is tested using a carefully calibrated fleet simulator of a large truckload carrier, using the offered loads from a major freight brokerage. We present the first model of an advance booking problem for truckload trucking where we have to make commitments to loads 1-14 days into the future. Although the load acceptance decision is approximated in the bidding process using a logistic curve with unknown parameters, actual load acceptance decisions are outputs of the simulator, which uses a stochastic lookahead policy that simulates fleet movements over a two week horizon for making advance commitments.  

The adaptation of the knowledge gradient policy for the high-dimensional contextual setting is new, and the simulations show that it outperforms a number of competitive learning policies.  The use of a stochastic lookahead model, implemented in a full fleet simulator, is also a first for truckload trucking, representing a very rigorous test of our bidding policy. We believe that the Est-Opt, Exploit and Mean-Price policies that we investigated represent policies that might reasonably be used in industry, and can thus be used as benchmarks.
We conclude that even though the carrier and shipper behavior is complex and highly stochastic, by carefully exploiting the value of information, we can significantly improve the total revenue and the number of contract acceptance.


\bibliographystyle{informs2014} 
\bibliography{article} 

\end{document}